\documentclass{article}

\PassOptionsToPackage{numbers, compress}{natbib}

\usepackage[final]{neurips_2025}

\usepackage[utf8]{inputenc} 
\usepackage[T1]{fontenc}    
\usepackage{url}            
\usepackage{hyperref}       
\usepackage{booktabs}       
\usepackage{amsfonts}       
\usepackage{nicefrac}       
\usepackage{microtype}      
\usepackage{xcolor}         

\usepackage{graphicx}
\usepackage{wrapfig}
\usepackage{enumitem}
\usepackage{amsmath}
\usepackage{multirow}
\usepackage{booktabs}
\usepackage[table]{xcolor}
\usepackage{makecell}
\usepackage{tgadventor}
\usepackage{tgheros}
\usepackage{microtype}
\usepackage{longtable}
\usepackage{pifont} 
\usepackage[scaled]{beramono}
\usepackage{caption}

\usepackage{xcolor} 

\newcommand{\ci}[1]{{\scriptsize \textit{\textcolor{gray!70}{±#1}}}}

\definecolor{pastelgreen}{RGB}{220,255,220}
\definecolor{pastelred}{RGB}{255,230,230}

\usepackage{inconsolata}

\newcommand{\rmsmall}{RM-8B}
\newcommand{\rmlarge}{RM-27B}
\newcommand{\method}{BLEUBERI}
\newcommand{\grporm}{\textsc{GRPO-RM}}
\newcommand{\veriscore}{\textsc{VeriScore}}
\newcommand{\fresh}{\textsc{FreshQA}}
\newcommand{\askh}{\textsc{AskHistorians}}
\newcommand{\longfact}{\textsc{LongFact}}

\newcommand{\cmark}{\textcolor{green!50!black}{\ding{51}}}
\newcommand{\xmark}{\textcolor{red!50!black}{\ding{55}}}
\newcommand{\claimtrue}[1]{\cellcolor{pastelgreen} #1 {\cmark}}
\newcommand{\claimfalse}[1]{\cellcolor{pastelred} #1 {\xmark}}

\title{\raisebox{-0.05em}{\includegraphics[height=0.8em]{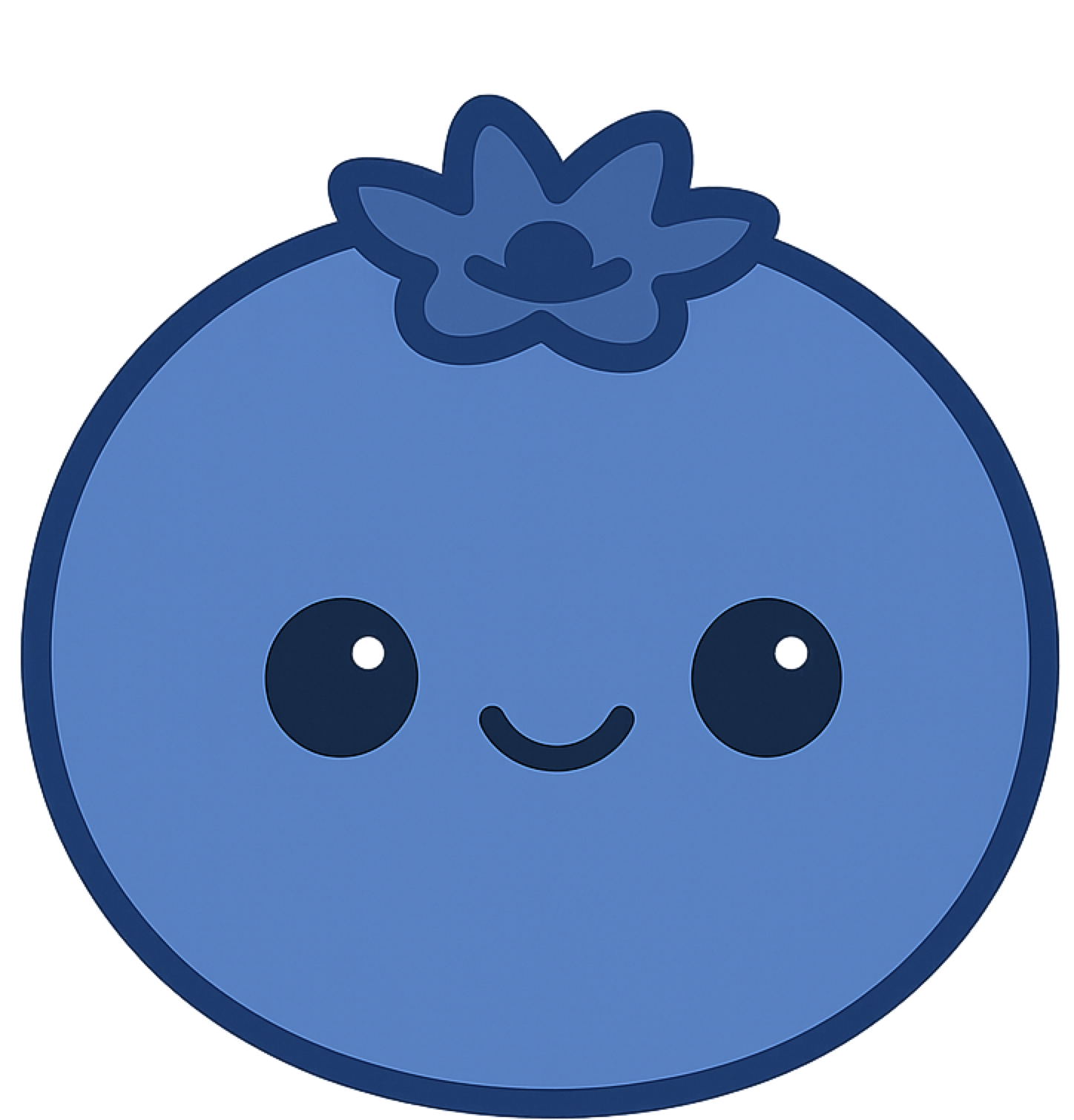}} \texttt{\method:} \\BLEU is a surprisingly effective reward\\for instruction following}

\author{%
  Yapei Chang \\
  University of Maryland, College Park
  \And
  Yekyung Kim \\
  University of Maryland, College Park
  \AND
  Michael Krumdick \\
  Kensho
  \And
  Amir Zadeh \\
  Lambda AI
  \And
  Chuan Li \\
  Lambda AI
  \And
  Chris Tanner \\
  Kensho
  \And
  Mohit Iyyer \\
  University of Maryland, College Park
}

\begin{document}

\maketitle

\begin{abstract} \label{sec:abstract}
   Reward models are central to aligning LLMs with human preferences, but they are costly to train, requiring large-scale human-labeled preference data and powerful pretrained LLM backbones. Meanwhile, the increasing availability of high-quality synthetic instruction-following datasets raises the question: can simpler, reference-based metrics serve as viable alternatives to reward models during RL-based alignment?
   In this paper, we show first that BLEU, a basic string-matching metric, surprisingly matches strong reward models in agreement with human preferences on general instruction-following datasets.
   Based on this insight, we develop \method,\footnote{\method\ stands for ``\textbf{BLEU}-\textbf{b}as\textbf{e}d \textbf{r}eward for \textbf{i}nstruction following.''} a method that first identifies challenging instructions and then applies Group Relative Policy Optimization (GRPO) using BLEU directly as the reward function.
   We demonstrate that \method-trained models are competitive with models trained via reward model-guided RL across four challenging instruction-following benchmarks and three different base language models.
   A human evaluation further supports that the quality of \method\ model outputs is on par with those from reward model-aligned models. Moreover, \method\ models generate outputs that are more factually grounded than competing methods.
   Overall, we show that given access to high-quality reference outputs (easily obtained via existing instruction-following datasets or synthetic data generation), string matching-based metrics are cheap yet effective proxies for reward models during alignment.
   We release our code and data at \href{https://github.com/lilakk/BLEUBERI}{\path{https://github.com/lilakk/BLEUBERI}}.\footnote{Correspondence to \texttt{yapeic@umd.edu}.}
\end{abstract}

\section{Introduction} \label{sec:introduction}

Modern LLM alignment often relies on reinforcement learning with a \emph{reward model} that guides the LLM to follow instructions according to human preferences \citep{ouyang2022training}. Reward models are expensive to train, requiring large-scale human preference data and powerful backbone models \citep{Ouyang2022TrainingLM, stiennon2022learningsummarizehumanfeedback, Touvron2023Llama2O}. Meanwhile, the emergence of high-quality instruction-following datasets (e.g., OpenHermes, Magpie, LIMA) has enabled cheaper  \emph{reference-based} alignment via supervised fine-tuning (SFT)  \citep{openhermes2024, xu2025magpie,zhou2023lima}. This contrast raises a natural question: \textbf{can we align language models with simple reference-based metrics in place of learned reward models?}

\paragraph{Challenges with reference-based rewards:} 
On the surface, replacing reward models with reference-based metrics seems problematic. Obtaining high-quality references for complex, open-ended tasks can be even more expensive than collecting preference judgments, and many instructions (e.g., creative writing) lack a single ground-truth answer.
Furthermore, aligning an LLM  requires balancing multiple criteria (e.g., helpfulness, harmlessness, factuality) \citep{bai2022traininghelpfulharmlessassistant}, that a single reference may not fully capture.
Finally, automatic metrics that score LLM responses against references rely on unreliable and gameable methods like $n$-gram matching or embedding similarity, which has historically discouraged their use in LLM alignment.

\vspace{-0.5em}
\paragraph{BLEU is surprisingly effective at modeling human preferences:}

Despite these limitations, we find that BLEU~\cite{bleu}, a simple string-matching metric long deemed inadequate for open-ended language generation \citep{xu2023critical, Evtikhiev_2023_outbleu, thai2022exploring, reiter-2018-structured, freitag-etal-2022-results, mathur-etal-2020-tangled, callison-burch-etal-2006-evaluating}, rivals large reward models in modeling human preferences.
In experiments on general instruction-following tasks in the LMSYS \texttt{chatbot\_arena\_conversations} dataset, BLEU with five synthetic references achieves almost the same agreement (\textbf{74.2\%}) with human preferences as a powerful 27B-parameter reward model (\textbf{75.6\%}).
We observe that reference quality is critical: references generated by powerful LLMs (e.g., Claude-3.7-Sonnet, GPT-4o) yield significantly higher agreement than those from weaker models.
Our analysis reveals that BLEU's strong alignment signal comes from rewarding properties like factuality and proper formatting that are critical to instruction following tasks. 

\vspace{-0.5em}
\paragraph{\method: directly using BLEU as a reward for RL-based alignment.}
Motivated by the unexpectedly high agreement between BLEU and human preferences, we propose \method, which uses RL to optimize BLEU on general instruction-following data. Previous efforts that used $n$-gram metrics like BLEU as rewards~\citep{ranzato2016sequenceleveltrainingrecurrent, Wu2018ASO, ive-etal-2021-exploring} faced obstacles such as unstable training \citep{bahdanau2017an, rennie2017} and degraded output quality \citep{liu-etal-2016-evaluate, khalifa2021a, korbak2022on}.
We revisit this line of work using modern LLMs within the paradigm of \emph{reinforcement learning with verifiable rewards} (RLVR) \citep{lambert2024tulu3, wang2025reinforcementlearningreasoninglarge}, which shows the effectiveness of simple, transparent rewards. \method\ treats BLEU as a verifiable reward for general instruction-following tasks, using group relative policy optimization (GRPO) \citep{shao2024deepseekmathpushinglimitsmathematical} to optimize a pretrained base LLM. We  apply GRPO with BLEU rewards on a subset of challenging instructions for which the base model's outputs initially have low BLEU.

\begin{wrapfigure}{r}{0.55\textwidth}
  \centering
  \includegraphics[width=0.55\textwidth]{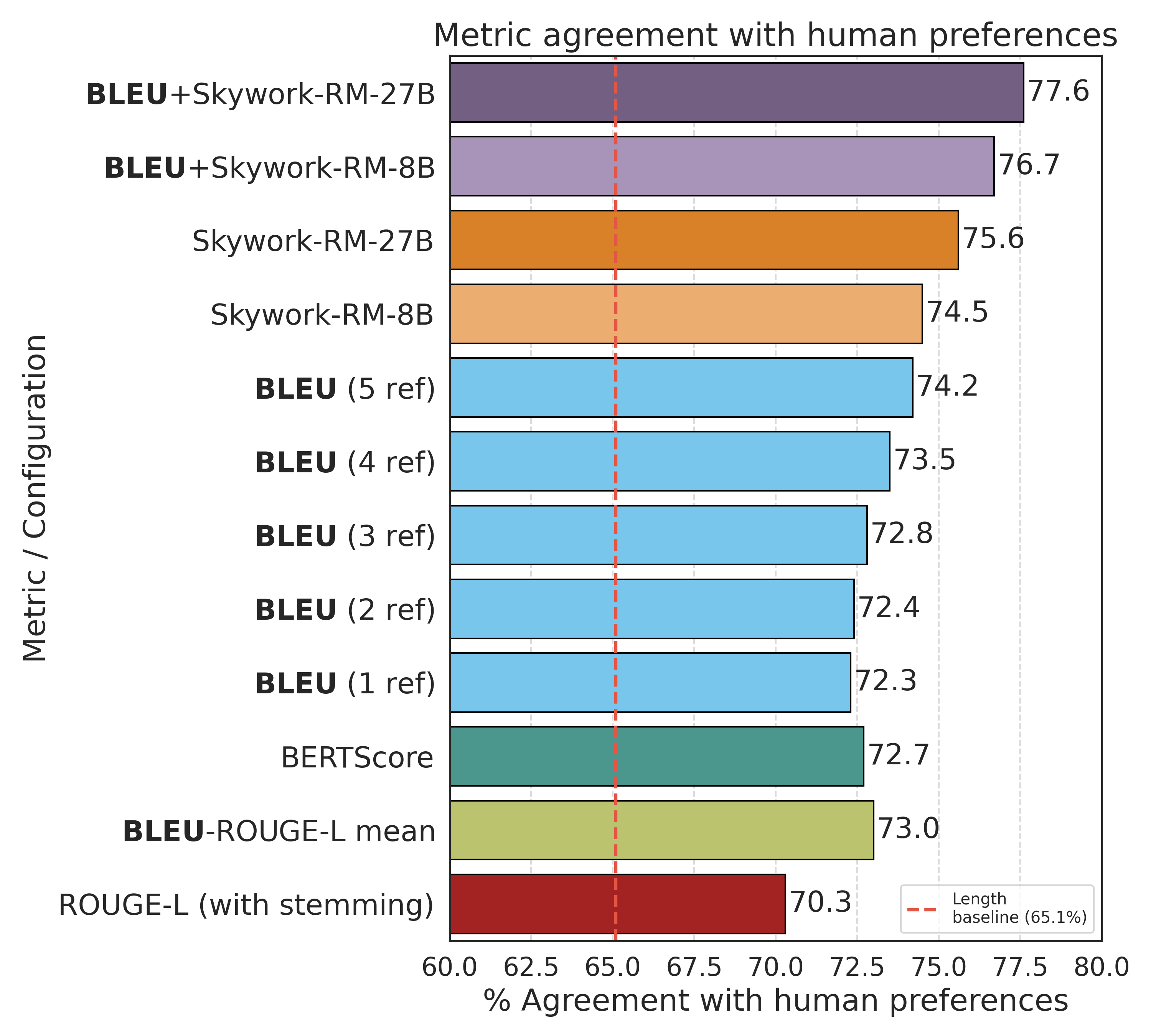}
  \caption{Human agreement rates for BLEU (with varying numbers of references), two reward models, and other reference-based metrics (with a single Claude reference). BLEU becomes more competitive with reward models as more references are provided, and combining BLEU with a reward model outperforms either alone.}
  \label{fig:arena-1k-multiref-and-other-metrics}
  \vspace{-0.5em}
\end{wrapfigure}

\vspace{-0.5em}
\paragraph{Strong instruction-following performance without a reward model:}
Across three different base models and four diverse instruction-following benchmarks, including ArenaHard\citep{li2024crowdsourced} and WildBench\cite{lin2025wildbench},  \method\ matches (and sometimes exceeds) the performance of reward model-guided RL and SFT according to both automatic and human evaluations. This is a striking result, given the simplicity and cost-effectiveness of \method\ compared to training and deploying large reward models for RLHF. Human evaluators find that \method-trained models are just as good as those from models aligned with reward models. Furthermore, \method\ models produce more factually-grounded responses than those aligned with either reward models or SFT. \textbf{Taken as a whole, ours is the first work to show that optimizing BLEU--far from overfitting to superficial $n$-gram matches--actually promotes helpful, factual, and well-formatted responses on general-domain instruction-following tasks.} Given access to high-quality synthetic references (readily available via existing datasets or generated from powerful LLMs), our \method\ method presents a novel alternative to alignment that entirely avoids training large, complex reward models.

\section{How well do simple reference-based metrics capture human preferences?} \label{sec:motivating}

In this section, we first investigate how well reference-based string matching metrics correlate with with human preference judgments on publicly available instruction-following datasets. To do so, we generate synthetic references for a subset of examples, and we find that BLEU is surprisingly competitive with state-of-the-art reward models in terms of agreement with human preferences. Moreover, BLEU's agreement with human preferences improves with more synthetic references (especially those from more powerful LLMs). 

\subsection{BLEU: an $n$-gram matching metric} \label{sec:bleu-definition}

BLEU (Bilingual Evaluation Understudy) \citep{bleu}
is a widely-used metric for machine translation evaluation. It measures the overlap between a predicted translation and \textbf{one or more} reference translations using modified $n$-gram precision ($n \in {1,2,3,4}$), combined with a brevity penalty (BP) to penalize overly short outputs:

$$
\mathrm{BLEU}=\mathrm{BP} \cdot \exp \left(\sum_{n=1}^N w_n \log p_n\right),\quad
\mathrm{BP} = 
\begin{cases}
1 & \text{if } c > r \\
\exp(1 - \frac{r}{c}) & \text{if } c \leq r
\end{cases}
$$

Here, $p_n$ is the modified $n$-gram precision, $w_n$ are their weights, and BP adjusts for length differences between the prediction ($c$) and the closest reference ($r$).\footnote{In our experiments, we use the huggingface implementation with \texttt{tokenizer\_13a}, and we apply smoothing to prevent zero scores for higher-order $n$-gram precisions when no matches are found. However, the SacreBLEU implementation is much faster and is therefore recommended for future use.}
When multiple references are available, BLEU computes $n$-gram matches and takes the maximum count across all references for each $n$-gram.

\subsection{How well does BLEU align with human preferences in single- and multi-reference setups?} \label{sec:arena-analysis}

To explore this question, we perform an analysis on the LMSYS \texttt{chatbot\_arena\_conversations} dataset~\citep{zheng2023judging}, which contains conversations evaluated by real users on Chatbot Arena.
Each instance includes an instruction, two model-generated outputs ($O_X$ and $O_Y$) and a human preference label.
We randomly select a subset of 900 instances from this dataset (more details in \S\ref{appendix-pref-analysis-getting-900}),\footnote{We do not evaluate on the full 33K dataset to reduce costs associated with collecting synthetic reference outputs from a variety of LLMs. In addition, later in our training experiments, we need to collect such synthetic references at a much larger scale, so we choose to minimize costs for this analysis.} then evaluate the following metrics on each pair of model outputs ($O_X$, $O_Y$):

\begin{itemize}[leftmargin=*]
    \item \textbf{Length baseline}: Prior work has found that humans tend to bias towards longer responses \citep{Chen2024HumansOL, Watts2024PARIKSHAAL, saito2023verbositybiaspreferencelabeling}, and that RLHF post-training may implicitly optimize for length over quality \citep{singhal2024longwaygoinvestigating}. To quantify the impact of output length, we implement a simple baseline that always prefers the longer output.
    
    \item \textbf{Reward models}: We use two strong reward models trained on well-curated preference data: Skywork-Reward-Gemma-2-27B-v0.2 (\textbf{\rmlarge}) and Skywork-Reward-Llama-3.1-8B-v0.2 (\textbf{\rmsmall}). These models assign scalar scores to responses based on a given instruction, without needing reference answers. They rank 4th and 11th, respectively, on RewardBench~\citep{RewardBench}.

    \item \textbf{BLEU}: Since the Chatbot Arena dataset lacks ground-truth responses, we construct a set of synthetic reference responses from a diverse set of LLMs.\footnote{These reference models include: Gemini-2.5-Pro~\citep{Gemini2_5-blog}, Claude-3.7-Sonnet~\citep{claude3_7-blog}, o4-mini~\citep{openai2025o3o4}, Deepseek-V3~\citep{deepseekai2025deepseekv3technicalreport}, Qwen-Max~\citep{qwen2025qwen25technicalreport}, GPT-4o~\citep{openai2024gpt4ocard}, Llama-3-8B-Instruct~\citep{llama3modelcard}, OLMo2-7B-Instruct~\citep{olmo20242olmo2furious}, and Qwen2.5-0.5B-Instruct~\citep{qwen2025qwen25technicalreport}. During the reference collection process, some closed-source models like Gemini would refuse to respond to certain prompts due to built-in safety filters. For our analysis, we only include prompts that received valid responses from all reference models. This results in a final set of 889 prompts.} For each instruction, we compute BLEU for $O_X$ and $O_Y$ using one or more references for this instruction. The response with the higher BLEU score is considered the winner. We evaluate both single- and multi-reference setups.

    \item \textbf{Other reference-based metrics}: We also evaluate ROUGE \citep{lin-2004-rouge} and BERTScore \citep{Zhang*2020BERTScore:}, two other popular reference-based metrics, later in this section.
\end{itemize}

\paragraph{BLEU agreement increases with more references.}
As established in \S\ref{sec:bleu-definition}, BLEU is a multi-reference metric. \autoref{fig:arena-1k-multiref-and-other-metrics} shows that increasing the number of references used by BLEU (up to five) improves its human agreement, reaching 74.2\%.
For comparison, the length baseline achieves 65.1\% agreement, whereas \rmsmall\ and \rmlarge\ respectively reach 76.7\% and 77.6\%.
To better understand BLEU's behavior, we show in \S\ref{appendix:pref-analysis-bleu-ablation} that both $n$-gram precision and the brevity penalty are necessary to achieve high human agreement.

\paragraph{Reference quality matters.}

Using a single reference and varying the reference model, we find that stronger models like Claude-3.7-Sonnet and GPT-4o yield over 72\% agreement, while weaker models like Qwen2.5-0.5B-Inst perform worse (60.9\%), even falling below the length baseline. 
More verbose models tend to yield lower agreement—e.g., Gemini-2.5-Pro scores only 69.5\% likely due to generating references 4.5x longer than outputs ($O_X$/$O_Y$), compared to Claude's 1.6x.
We observe a strong negative Pearson correlation (-0.78) between the absolute difference in length (between the reference and the outputs $O_X$/$O_Y$) and BLEU's agreement with human preferences. We hypothesize that this is because the number of unmatched $n$-grams increases as the responses get longer, washing out any distinguishing signal. See \S\ref{sec:appendix-pref-analysis-len-impact} for more analysis on length effects.

\begin{wrapfigure}{r}{0.42\textwidth}
  \centering
  \includegraphics[width=0.42\textwidth]{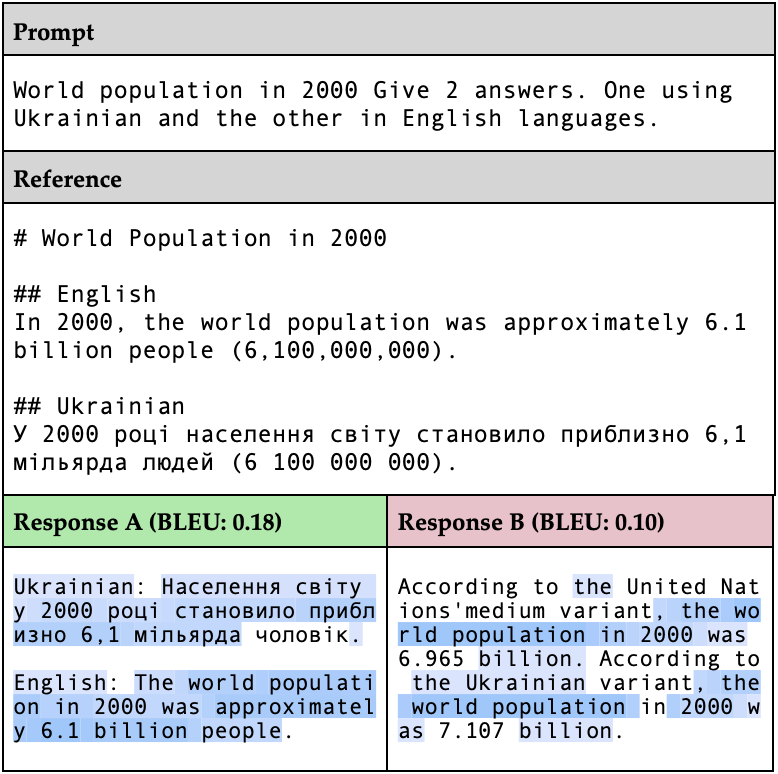}
  \caption{The highlighted matching $n$-grams in this example show that BLEU can capture correct instruction-following behavior as well as the factuality of the response.}
  \label{fig:bleu_example1}
  \vspace{-1.5em}
\end{wrapfigure}
\paragraph{Other reference-based metrics also exhibit high agreement.}
We also evaluate ROUGE \citep{lin-2004-rouge}, which measures $n$-gram recall, and BERTScore \citep{Zhang*2020BERTScore:}, which computes contextual embedding similarity. We also evaluate BLEU+RM, a combined metric integrating BLEU with reward models via z-score normalization and averaging.\footnote{We apply z-score standardization to both scores across all examples, then average them to compute a combined score per example.}
As shown in \autoref{fig:arena-1k-multiref-and-other-metrics}, ROUGE-L (with stemming),\footnote{Without stemming, we observe a drop of 1.8 in agreement with human preferences.} BERTScore, and BLEU-ROUGE harmonic mean all exhibit human agreement comparable to BLEU.
Notably, BLEU+RM (with a single Claude reference) reaches higher agreement than either BLEU or the reward model alone, suggesting that they each focus on different aspects of the response.
While all of these other rewards are promising to explore, we focus on BLEU for the remainder of this paper due to its simplicity and our limited computational resources.

\paragraph{Types of $n$-grams that contribute to BLEU's agreement:}
When BLEU makes a pairwise judgment that aligns with human preferences, which $n$-grams contribute to that decision?
We show one such example in \autoref{fig:bleu_example1}.
Here, the presence of \texttt{Ukrainian} and \texttt{English} headers, along with the value \texttt{6.1}, contribute to its higher BLEU score, showing that BLEU can effectively capture  both format and factuality (additional examples in \S\ref{sec:appendix-pref-analysis-ngram-attribution}).

\subsection{Domain-specific agreement analysis} \label{sec:arena-by-domain}
To assess BLEU’s effectiveness across domains, we analyze 18,219 filtered examples from the Chatbot Arena dataset using Claude-3.7-Sonnet as the reference model.
Examples are grouped into six domains: \textit{QA}, \textit{Code}, \textit{Writing}, \textit{Math/Reasoning}, \textit{Multilinguality}, and \textit{Planning}. \autoref{fig:agreement_category_all} shows BLEU aligns best with reward models in \textit{Writing} and \textit{Code}, but performs poorly in \textit{Math/Reasoning}, likely due to surface-level variation in mathematically equivalent expressions.
Limitations may also stem from \texttt{tokenizer\_13a}'s handling of equations and non-whitespace-delimited languages. Full figure and additional details on this analysis are in \S\ref{sec:app-diff-domain}.

\section{\method: aligning language models to follow instructions with BLEU} \label{sec:training}

Having established that BLEU agrees with human judgments at a similar rate to powerful reward models, we turn to actually using BLEU as a reward during LLM alignment. 
Below, we describe our experiment setups and results.
In total, all our training experiments (including main and exploratory runs) cost \textbf{1357} Nvidia GH200 hours.
Across four benchmarks, models trained with \method\ perform comparably to those trained with SFT or GRPO-RM.
To our knowledge, this is the first work to demonstrate that reinforcement learning using BLEU directly as the reward can be surprisingly effective for general instruction following.

\paragraph{Training methods for alignment:}

Besides standard SFT, we also train models using Group Relative Policy Optimization (GRPO) \citep{shao2024deepseekmathpushinglimitsmathematical}, which fine‑tunes a language model by sampling $K$ candidate responses ${y_k}$ for a prompt $x$, scoring them with $R(y_k,x)$, and computing the group‑normalized advantage.
We instantiate the reward function \(R(y_k, x)\) in two ways:

\begin{enumerate}[leftmargin=*, label=\arabic*.]
    \item \textbf{GRPO-RM}: \( R(y_k, x) = R_{\text{RM}}(y_k, x) \), where the reward is provided by \rmsmall.\footnote{We do not train with \rmlarge\ due to computational constraints. As shown in \S\ref{sec:motivating}, both \rmlarge\ and \rmsmall\ achieve comparable agreement with human preferences.} While reward models are often used in methods like PPO \citep{schulman2017proximalpolicyoptimizationalgorithms}, they can also be directly used in GRPO to score and rank outputs within a group \citep{shao2024deepseekmathpushinglimitsmathematical, yang2024qwen25mathtechnicalreportmathematical}.
    
    \item \textbf{\method}: \( R(y_k, x) = \text{BLEU}(y_k, \text{Ref}(x)) \), where the reward is the BLEU score computed against one or more reference responses for $x$.
\end{enumerate}

We use GRPO with a reward model to enable a controllable comparison to \method, which uses GRPO with a BLEU reward. GRPO has been shown to be just as effective as other algorithms like PPO \citep{schulman2017proximalpolicyoptimizationalgorithms} when used with a reward model \citep{jiao-etal-2024-learning, wang2025ragenunderstandingselfevolutionllm}.
Scoring outputs with BLEU can be up to 48 times faster than using \rmsmall\ (more in \S\ref{appendix-training-runtimes}).
Although GRPO has been widely applied to reasoning-intensive tasks \citep{deepseekai2025deepseekr1incentivizingreasoningcapability, xu2025phi4minireasoningexploringlimitssmall}, our experiments enforcing reasoning behavior did not yield strong results (more in \ref{appendix:training-reasoning}).

\subsection{Training data} \label{sec:data-selection}

In preliminary experiments, we find that while \method\ can work with randomly-selected data, it performs best when trained on data with low initial rewards. Specifically, we first construct a data pool of prompts, run the base model on each, and compute the BLEU score of its outputs against the reference responses. Prompts with the lowest BLEU scores are treated as ``hard'' examples.
Full experimental details on the data difficulty ablation are provided in \S\ref{appendix:training-difficulty-ablation}.
While the selection of hard data requires collecting references for a larger pool, obtaining them (via existing datasets or LLMs) is still generally cheaper than collecting human preference data and training large reward models.

\paragraph{Creating a data pool:}
We draw from the Tulu3 SFT mixture \citep{lambert2024tulu3}, which contains 939K examples across 18 data sources covering diverse tasks.
Motivated by the strong human agreement of BLEU on writing tasks from \S\ref{sec:arena-by-domain}, we filter and sample 50K prompts to form our final data pool, where the majority of examples are related to writing. See more details in \S\ref{appendix:training-tulu3}.

\paragraph{Training on hard examples:}
To ensure consistency in our main experiments, we train all methods on the 5,000 hardest examples as ranked by BLEU.
While this strategy might seem biased against GRPO-RM, we also train GRPO-RM on hard negatives selected by \rmsmall\ in \S\ref{appendix:training-rm-selected} and show similar performance, indicating that hard negatives do not matter as much when using a reward model.
Concurrent work also highlights the benefits of training on difficult data: the GRPO pipeline in the recently released Qwen3 report trains exclusively on 4,000 samples selected to be as challenging as possible \citep{qwen3technical2025}.

\paragraph{Collecting references for \method:}
Unlike the Chatbot Arena dataset used in \S\ref{sec:motivating}, Tulu3 includes ground-truth responses for each instruction, which we refer to as \textit{\textbf{Tulu3 references}}.
Within our 50K data pool, \textbf{45.2\%} of the \textit{Tulu3 references} are human-written responses sourced from previous datasets, while the remaining half are synthetic outputs from LLMs like ChatGPT.
In addition, to evaluate the effectiveness of using completely synthetic references from different LLMs, we collect outputs from Claude-3.7-Sonnet, Gemini-2.5-Pro, o4-mini, Deepseek-V3, and Llama-3.1-8B-Instruct (more details in \S\ref{appendix:training-ref-ablation}), and refer to them as \textit{\textbf{additional synthetic references}}.
In our main experiments (\autoref{tab:main-result}), we use \textit{Tulu3 references} for hard data selection and training, and discuss results using \textit{additional synthetic references} later in \S\ref{sec:training-results}.

\subsection{Training and evaluation setup} \label{sec:training-setup}

\paragraph{Base models and hyperparameters:}
We conduct our experiments using four base models: Qwen2.5-14B, Qwen2.5-7B, Qwen2.5-3B, and Llama-3.1-8B.\footnote{We do not use the instruct (officially post-trained) versions of these models, as they have already undergone extensive post-training optimization for instruction following.} Compared to Qwen, Llama-3.1-8B is a significantly weaker base model, having likely seen minimal or no instruction-following data during pretraining. To address this limitation, we train Llama-3.1-8B on the 50K data pool for one epoch to equip it with a basic ability to handle chat-style prompts. We refer to this SFT-initialized version of Llama as the ``Llama base model'' in the remainder of this work.
This setup aligns with prior observations that effective GRPO training requires a sufficiently capable base model \citep{liu2025understandingr1zeroliketrainingcritical}.
For details about GRPO on Qwen after SFT initialization, see \S\ref{appendix:training-sft-grop-qwen}.
For all methods, we train for one full epoch on the BLEU-selected 5K data.\footnote{For SFT, we use a learning rate of 5e-6 with a global batch size of 32 and set max tokens (covering both input and output) to 1024. For GRPO, we set the learning rate to 1e-6, group size of 8, max prompt length and max generation length to 512 tokens, and maintain the same global batch size of 32---meaning each batch consists of 8 generations for each of 4 unique prompts. All training runs are performed on single GH200 GPUs using TRL with DeepSpeed-ZeRO3 (\url{https://www.deepspeed.ai/2021/03/07/zero3-offload.html}).}
Note that GRPO and SFT use fundamentally different optimization strategies, so it is hard to ensure a completely fair comparison. More discussion on this and results for multi-epoch SFT training are in \S\ref{appendix-training-multiepoch}.

\paragraph{Benchmarks:}
We evaluate our models using four benchmarks:
\textbf{MT-Bench} \citep{zheng2023judging}, a set of 80 manually curated, high-quality multi-turn questions;
\textbf{ArenaHard v1 and v2} \citep{li2024crowdsourced}, two distinct sets of 500 challenging prompts drawn from real-world user queries. ArenaHard v2, released in April 2025, contains an updated collection with more difficult prompts than v1;
and \textbf{WildBench} \citep{lin2025wildbench}, comprising 1,024 complex real-world queries.
Inference on these benchmarks are run using vLLM \citep{kwon2023efficient} with greedy coding.
We also report 95\% confidence intervals computed from bootstrapping.
More details on running each benchmark are in \S\ref{sec:appendix-training-benchmark-setup}.
All benchmarks are evaluated using the LLM-as-a-judge framework. To balance cost and performance, we select \texttt{gpt-4.1-mini} as the judge for all benchmarks.

\begin{table}[t!]
\caption{Results on four instruction-following benchmarks with 95\% confidence intervals. For each model, the ``\textit{Base}'' row represents its pretrained checkpoint (for Llama-3.1-8B, this is our SFT-initialized model described in \S\ref{sec:training-setup}), while the \textit{Instruct} row is the official post-trained checkpoint. Despite the limitations of $n$-gram matching, \method\ remains competitive with both SFT and GRPO-RM across all models and benchmarks, as indicated by the closely aligned scores and frequently overlapping confidence intervals.}
\label{tab:main-result}
\centering
\scriptsize
\resizebox{\textwidth}{!}{
\begin{tabular}{l l c c c c c}
\toprule
\textbf{Model} & \textbf{Variant} & \textbf{MT-Bench} & \textbf{ArenaHard v1} & \textbf{ArenaHard v2} & \textbf{WildBench} & \textbf{Average} \\
\midrule

\multirow{5}{*}{Qwen2.5-14B}
  & \textit{Base} & 66.3 \ci{3.8} & 20.9 \ci{2.1} & 8.0 \ci{1.0} & 57.2 \ci{1.4} & 38.1 \\
  & \hspace{0.7em}$\hookrightarrow$ SFT & 70.7 \ci{3.9} & 29.0 \ci{2.5} & 13.6 \ci{1.3} & 63.5 \ci{1.3} & 44.2 \\
  & \hspace{0.7em}$\hookrightarrow$ GRPO-RM & 78.6 \ci{4.0} & 36.0 \ci{2.5} & 16.3 \ci{1.4} & 68.4 \ci{1.2} & 50.5 \\
  & \cellcolor{blue!10}\hspace{0.7em}$\hookrightarrow$\raisebox{-0.05em}{\includegraphics[height=0.8em]{assets/bleuberry.png}}\method & \cellcolor{blue!10}74.2 \ci{3.8} & \cellcolor{blue!10}32.8 \ci{2.2} & \cellcolor{blue!10}16.5 \ci{1.4} & \cellcolor{blue!10}66.7 \ci{1.2} & \cellcolor{blue!10}47.6 \\
  & \textit{Instruct} & 84.6 \ci{3.0} & 58.5 \ci{2.7} & 29.7 \ci{1.8} & 75.7 \ci{0.9} & 62.1 \\
\midrule

\multirow{5}{*}{Qwen2.5-7B}
  & \textit{Base} & 63.5 \ci{4.1} & 16.2 \ci{1.6} & 5.6 \ci{0.8} & 51.8 \ci{1.6} & 34.3 \\
  & \hspace{0.7em}$\hookrightarrow$ SFT & 67.5 \ci{3.9} & 22.1 \ci{2.1} & 9.9 \ci{1.1} & 60.5 \ci{1.3} & 40.0 \\
  & \hspace{0.7em}$\hookrightarrow$ GRPO-RM & 76.9 \ci{3.5} & 29.8 \ci{2.2} & 12.2 \ci{1.2} & 64.8 \ci{1.1} & 45.9 \\
  & \cellcolor{blue!10}\hspace{0.7em}$\hookrightarrow$\raisebox{-0.05em}{\includegraphics[height=0.8em]{assets/bleuberry.png}}\method & \cellcolor{blue!10}70.8 \ci{3.7} & \cellcolor{blue!10}29.3 \ci{2.2} & \cellcolor{blue!10}12.9 \ci{1.3} & \cellcolor{blue!10}65.4 \ci{1.3} & \cellcolor{blue!10}44.6 \\
  & \textit{Instruct} & 78.8 \ci{3.4} & 37.9 \ci{2.7} & 16.8 \ci{1.4} & 71.0 \ci{1.1} & 51.1 \\
\midrule

\multirow{5}{*}{Qwen2.5-3B}
  & \textit{Base} & 61.1 \ci{3.8} & 7.0 \ci{1.0} & 3.1 \ci{0.6} & 49.0 \ci{1.3} & 30.1 \\
  & \hspace{0.7em}$\hookrightarrow$ SFT & 59.6 \ci{4.2} & 9.8 \ci{1.1} & 4.0 \ci{0.7} & 55.6 \ci{1.4} & 32.2 \\
  & \hspace{0.7em}$\hookrightarrow$ GRPO-RM & 67.8 \ci{4.1} & 12.8 \ci{1.4} & 5.1 \ci{0.8} & 59.2 \ci{1.2} & 36.2 \\
  & \cellcolor{blue!10}\hspace{0.7em}$\hookrightarrow$\raisebox{-0.05em}{\includegraphics[height=0.8em]{assets/bleuberry.png}}\method & \cellcolor{blue!10}64.6 \ci{4.0} & \cellcolor{blue!10}11.0 \ci{1.3} & \cellcolor{blue!10}3.7 \ci{0.6} & \cellcolor{blue!10}56.1 \ci{1.2} & \cellcolor{blue!10}33.8 \\
  & \textit{Instruct} & 70.0 \ci{3.9} & 16.8 \ci{1.5} & 6.7 \ci{0.9} & 63.0 \ci{1.2} & 39.7 \\
\midrule

\multirow{5}{*}{\makecell[l]{Llama-3.1-8B \\ \scriptsize (SFT init.)}}
  & \textit{Base} & 52.9 \ci{4.0} & 6.8 \ci{1.1} & 2.1 \ci{0.4} & 55.0 \ci{1.3} & 29.2 \\
  & \hspace{0.7em}$\hookrightarrow$ SFT & 56.7 \ci{4.5} & 12.6 \ci{1.6} & 3.0 \ci{0.6} & 60.3 \ci{1.3} & 33.2 \\
  & \hspace{0.7em}$\hookrightarrow$ GRPO-RM & 57.2 \ci{4.3} & 9.7 \ci{1.4} & 1.6 \ci{0.5} & 57.8 \ci{1.4} & 31.6 \\
  & \cellcolor{blue!10}\hspace{0.7em}$\hookrightarrow$\raisebox{-0.05em}{\includegraphics[height=0.8em]{assets/bleuberry.png}}\method & \cellcolor{blue!10}56.9 \ci{4.3} & \cellcolor{blue!10}10.1 \ci{1.5} & \cellcolor{blue!10}2.3 \ci{0.5} & \cellcolor{blue!10}59.5 \ci{1.2} & \cellcolor{blue!10}32.2 \\
  & \textit{Instruct} & 65.8 \ci{3.8} & 24.9 \ci{2.4} & 5.8 \ci{0.9} & 64.2 \ci{1.2} & 40.2 \\
\bottomrule

\end{tabular}
}
\end{table}

\subsection{Experimental results}  \label{sec:training-results}

\paragraph{\method\ performs on par with GRPO-RM and SFT across all benchmarks.}
Across four base models and four benchmarks, \method\ achieves performance comparable to GRPO-RM (\autoref{tab:main-result}), as indicated by their closely matched scores and frequently overlapping 95\% confidence intervals. This aligns with the human agreement analysis presented in the previous section.

\paragraph{\method\ does not compromise creativity.}
One potential concern with \method's single-reference optimization is that it could lower performance on open-ended or creative tasks where a single instruction has many valid responses. To evaluate this, we examine performance on the WildBench Creative Tasks split. The results show that Qwen2.5-7B models trained with \method\ (66.7), GRPO-RM (67.2), and SFT (60.4) perform similarly, indicating that BLEU-based optimization does not hinder creative capabilities. See detailed results in \S\ref{sec:app-creative}.

\paragraph{\method\ with fully synthetic references also shows strong performance.}
In \S\ref{sec:arena-analysis}, we analyze how well BLEU aligns with human judgments in a single-reference setting, varying the reference model. To investigate whether stronger BLEU-human agreement translates into better training outcomes, we train Qwen2.5-7B using the \textit{additional synthetic references} collected earlier. We observe a Pearson correlation of 0.34 between each reference model’s human agreement and the resulting trained model's performance, indicating a moderately strong relationship. Among these, Claude and o4-mini produce the best results, matching the performance of GRPO-RM.
We also experiment with a 5-reference training setup and find that the resulting model outperforms most of the models trained using only a single reference from the same set.
Further experimental details and discussion are provided in \S\ref{appendix:training-ref-ablation}.

\paragraph{Training with reward functions beyond BLEU:}
Due to computational constraints, our experiments primarily focus on BLEU as the reward function. Nevertheless, we find that GRPO training with alternative rewards---such as BERTScore and BLEU-ROUGE harmonic mean---can achieve performance comparable to BLEU.
We provide details on our experiments with these alternative rewards in \S\ref{appendix:training-other-rewards}, and hope our findings can encourage more comprehensive explorations of other reward functions in future work.

\paragraph{Random rewards do not yield meaningful gains on general instruction following.}
\citet{shao2025spurious} show that random rewards improve math reasoning in Qwen2.5-Math models \citep{yang2024qwen25mathtechnicalreportmathematical}, but not in Llama \citep{llama3modelcard} or OLMo \citep{olmo20242olmo2furious}, by eliciting latent reasoning strategies. We test whether this signal also aids general instruction following by training Qwen2.5-7B and Qwen2.5-3B with GRPO using a random reward defined as \( R(y_k, x) \sim \text{Uniform}(0, 1) \), and report results in \S\ref{appendix-training-random-rewards}.
Unlike \citet{shao2025spurious}, who find large gains in math reasoning, we observe no improvement on instruction-following benchmarks as we see no meaningful change for Qwen2.5-7B and degraded performance for Qwen2.5-3B. This suggests that \method’s benefits are domain-specific, and extending RLVR beyond math remains an open challenge.
\section{Analysis and human evaluation of model outputs} \label{sec:results-and-analysis}

In this section, we extend beyond benchmark numbers and look into various qualitative properties of model-generated outputs. Our qualitative observation is that the GRPO-trained model notably follows instructions, whereas the base model often fails to do so. We also find that SFT-trained models tend to generate more verbose and repetitive text compared to \method, while GRPO-trained models more frequently use markdown formatting (see \S\ref{appendix-training-qual-stats} for detailed example and results).
Outputs generated by \method\ are also more factually accurate than GRPO-RM and SFT, while a human evaluation shows that \method's outputs are rated similarly to those of GRPO-RM.

\subsection{Surface-level qualitative characteristics} \label{sec:qual-analysis}

In \autoref{tab:output-qual-stats-overall}, we show that SFT models tend to produce more verbose and repetitive outputs. 
GRPO-RM models exhibit a slightly higher refusal rate (i.e., where the model refuses to follow the user instruction). Both \method\ and GRPO-RM models use markdown formatting more frequently than SFT models, with GRPO-RM showing the highest usage. In \S\ref{appendix-training-qual-stats}, we explain how these statistics are computed and provide example outputs.

\begin{wrapfigure}{r}{0.45\textwidth}
  \centering
  \includegraphics[width=0.45\textwidth]{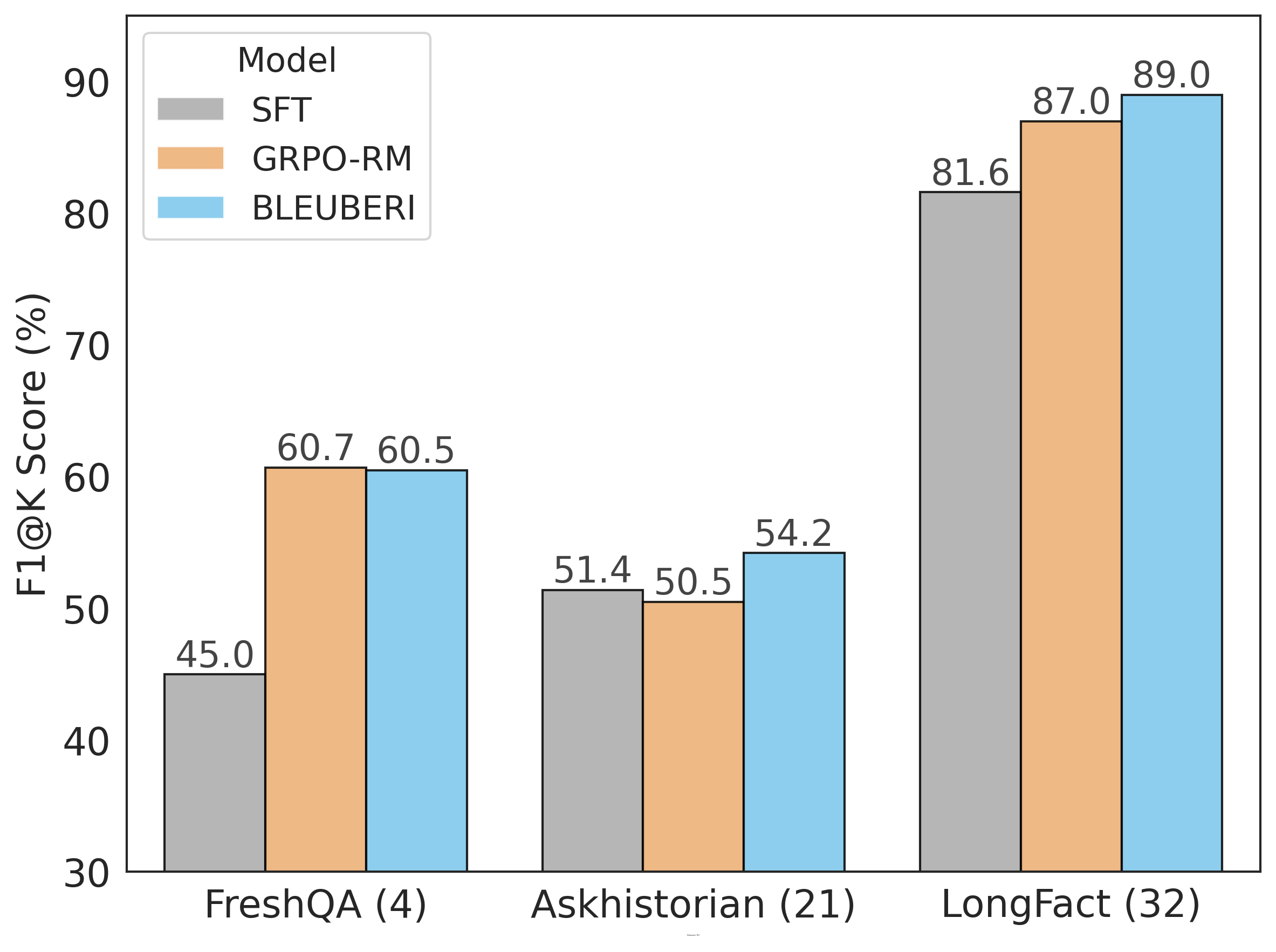}
  \caption{Factuality results for trained Qwen2.5-7B models across three QA datasets evaluated using \textsc{VeriScore}~\citep{song-etal-2024-veriscore}. The $K$ values (in parentheses on the
  x-axis) used for each dataset follow the original paper.}
  \label{fig:factuality_veriscore}
  \vspace{-0.7em}
\end{wrapfigure}

\paragraph{Qwen models trained with GRPO frequently use affirmative openers.}

\method-trained Qwen models frequently begin responses with \texttt{Certainly!''} (51.6\% for Qwen2.5-3B, 27.1\% for Qwen2.5-7B), while GRPO-RM models often use \texttt{Sure!''} (70.6\% and 35.2\%). These phrases are rare in Qwen base models and all Llama variants.\footnote{SFT models often begin with \texttt{To...} (e.g., \texttt{To address your query...}) in $\sim$20–25\% of responses, similar to base models, suggesting this is not learned during SFT.}
The frequent use of \texttt{``Certainly!''} is unlikely to be reward hacking, as the phrase appears in under 1\% of the Tulu3 references used in our main training experiments. A more plausible explanation is that GRPO amplifies subtle biases latent in Qwen's pretraining, even if absent in base model outputs. This tendency has been observed previously:
for instance, the system prompt for a previous version of Claude explicitly states: \texttt{Specifically, Claude avoids starting responses with the word ``Certainly'' in any way.} \footnote{\url{https://docs.anthropic.com/en/release-notes/system-prompts}, specifically the system prompt for Claude-3.5-Sonnet, July 12th, 2024.}
See more details on this analysis in \S\ref{appendix:analysis-affirmatives}.

\subsection{Factuality of model outputs}

\begin{figure}[htbp]
  \centering
  \includegraphics[width=1.0\textwidth]{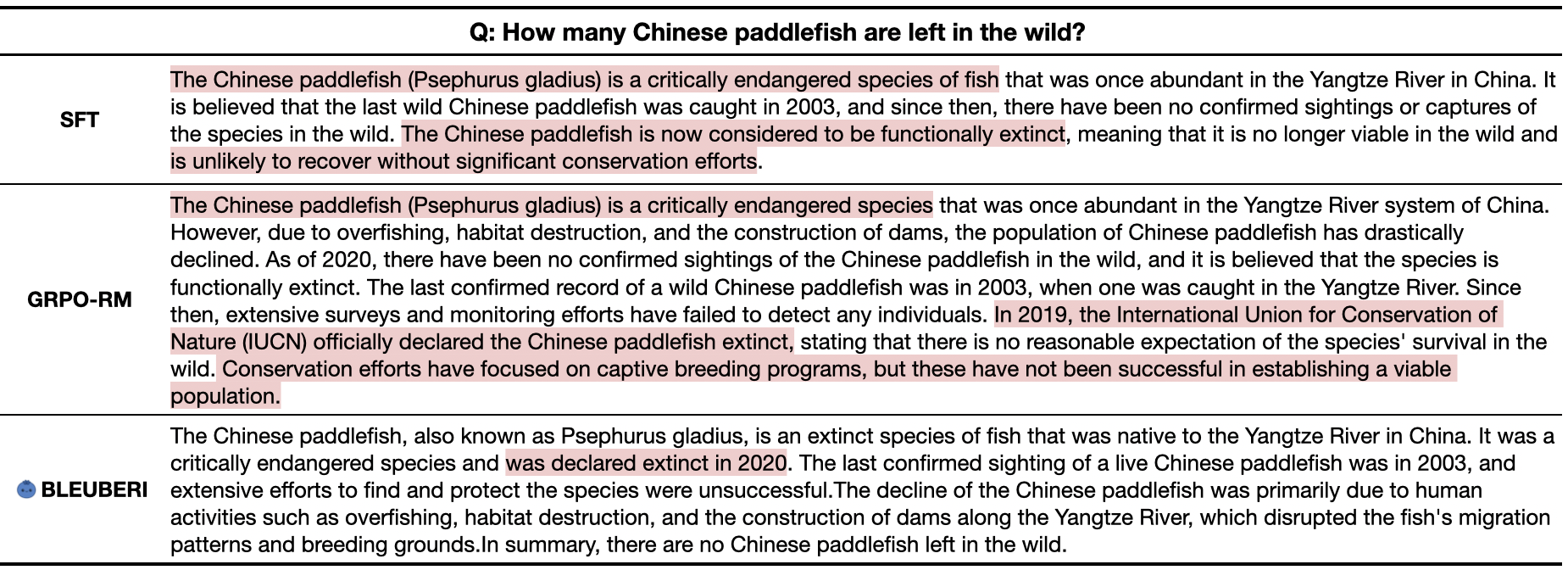}
\caption{An example instruction from FreshQA, where red highlights indicate factually incorrect claims. For this instruction, \method\ produces a more factually precise output than GRPO-RM.}
  \label{fig:factuality_exmaple}
\end{figure}

While BLEU has often been criticized for its weak correlation with factual accuracy due to its lack of sensitivity to semantic content \citep{kanerva-etal-2019-template, xu-etal-2019-asking, wei2019conductingbettervalidationstudies, shailya2025lextevaluatingtrustworthinessnatural}, recent studies have explored more effective ways to leverage BLEU for factual text generation \citep{Thulke2023GroundedDialog, Haldar2024code, mostafazadeh-etal-2020-glucose}.
Building on this line of work, we evaluate the factual accuracy of variants of Qwen2.5-7B models, using \veriscore\ \citep{song-etal-2024-veriscore}---an automatic metric that evaluates factuality in long-form generation tasks.\footnote{\veriscore\ measures the factuality within the response itself at a claim level; It does not necessarily indicate whether the response correctly answers the question.}
We use three datasets covering different domains and report F1@$K$ as defined in \veriscore\, where recall is relative to a median number of claims $K$ for a given dataset.
\autoref{fig:factuality_veriscore} shows that \method\ consistently outperforms GRPO-RM and SFT on \longfact\ and \askh\, and performs on par with \grporm\ on \fresh. These results suggest that optimizing for BLEU using RL can guide models toward more factually grounded generations.
See example outputs in \autoref{fig:factuality_exmaple} and \S~\ref{appendix:factuality} for more evaluation details.

\subsection{Human evaluation of \method\ and GRPO-RM model outputs}

Although \method\ models performs competitively with GRPO-RM models on benchmarks, a valid concern is that its reliance on $n$-gram matching could lead to reward hacking, as such matches are relatively easy to exploit.
To assess whether \method\ produces outputs that align with human preferences, we conduct a human evaluation comparing outputs from Qwen2.5-7B models trained with GRPO-RM and \method.
We sample 120 examples (30 from each benchmark) and asked two annotators to compare the outputs from the BLEU-trained and RM-trained models, denoted $O_B$ and $O_R$ respectively.\footnote{These annotators are co-authors of this paper, but they did not participate in setting up the annotation task and had no prior exposure to any model outputs.}
To prevent implicit bias, we remove affirmative openers for each model if they exist in the outputs (see \S\ref{sec:qual-analysis}). For each pair, we anonymize the model identity and randomize their order. In addition to direct preferences, we also allow a ``tie'' option.

\paragraph{Humans judge \method\ outputs to be equally as good as those from GRPO-RM.}
We compute a soft preference rate, defined as the proportion of times an annotator judges $O_B$ to be at least as good as $O_R$.\footnote{Soft preference rate = $O_B$ wins + ties.}
While this metric provides a coarse measure of overall favorability toward $O_B$, it should not be interpreted as equivalent to a strict preference rate, since ties do not necessarily indicate endorsement.
The results are visualized in \autoref{fig:human-preferences}. We observe a soft preference for $O_B$ in 67.5\% of Annotator 1's judgments and 52.5\% of Annotator 2's. Out of the 63 cases where both annotators express a clear preference (i.e., not a tie), the Cohen’s Kappa agreement is 0.34, indicating fair agreement.\footnote{Our annotation task required subjective judgments on highly similar, high-quality responses to complex prompts from four benchmarks (MT-Bench, ArenaHard v1/v2, WildBench). These often involved technical or domain-specific content, making clear preferences difficult. The frequent use of the ``tie'' option by one annotator further suggests the closeness in output quality.} These results suggest that human evaluators generally view \method\ outputs as roughly comparable to those of GRPO-RM. More details in \S\ref{sec:appendix-human-eval}.

\section{Related work} \label{sec:related-work}

\paragraph{Prior approaches to training with BLEU optimization:}
A substantial body of prior work has explored sequence-level reinforcement learning for tasks such as machine translation, using BLEU as a reward signal. One line of research centers on Minimum Risk Training \citep{och-2003-minimum, smith-eisner-2006-minimum, zens-etal-2007-systematic, li-eisner-2009-first, kiegeland-kreutzer-2021-revisiting}, which directly minimizes expected task-specific loss, enabling optimization of non-differentiable metrics like BLEU. Another line, grounded in policy gradient methods such as REINFORCE \citep{reinforce-1992}, faces well-known challenges including high-variance gradient estimates \citep{bahdanau2017an, rennie2017} and degraded output quality \citep{liu-etal-2016-evaluate, khalifa2021a, korbak2022on}. To mitigate exposure bias, \citet{ranzato2016sequenceleveltrainingrecurrent} proposed the MIXER algorithm, which also directly optimizes BLEU through a mixed objective.
Despite their prevalence, BLEU and other $n$-gram-based metrics have been shown to correlate poorly with human judgments across multiple domains, including machine translation \citep{callison-burch-etal-2006-evaluating, reiter-2018-structured, mathur-etal-2020-tangled, freitag-etal-2022-results, wieting-etal-2019-beyond}, summarization \citep{kryscinski-etal-2019-neural}, code generation \citep{bleu-for-code}, and question answering \citep{xu-etal-2023-critical, wang2020askingansweringquestionsevaluate}. In this context, our work is the first to investigate the use of BLEU as a training signal for general instruction following with modern LLMs.

\paragraph{RLVR with simple metrics in other domains:}
Recent work has found that in mathematical reasoning tasks, reinforcement learning with simple rule-based rewards can be surprisingly effective even without the use of reward models \citep{deepseekai2025deepseekr1incentivizingreasoningcapability, wang2025reinforcementlearningreasoninglarge, yu2025dapoopensourcellmreinforcement, xu2025phi4minireasoningexploringlimitssmall}.
Beyond math, similar efforts have extended to other domains such as story generation \citep{gurung2025learningreasonlongformstory}, visual perception \citep{liu2025visualrftvisualreinforcementfinetuning}, and medical reasoning \citep{zhang2025medrlvremergingmedicalreasoning}.
Of particular relevance to our work is that of \citet{lambert2024tulu3}, which introduces the term ``RLVR'' and investigates verifiable rewards for synthetically-constrained tasks; we build on this work by using BLEU as a form of verifiable reward for general instruction following.

\section{Conclusion}

In this paper, we revisit reference-based metrics for aligning LLMs and demonstrate that BLEU—a simple $n$-gram overlap metric—correlates surprisingly well with human preference judgments on general instruction-following tasks. While earlier attempts to directly optimize BLEU with RL ran into challenges with unstable training and generated artifacts, our results show that BLEU is indeed a practical reward signal for modern alignment. Our \method\ approach, which optimizes BLEU on open-weight language models using the recently-developed GRPO algorithm,  achieves performance on par with reward model-guided RL across diverse instruction-following benchmarks and model scales. Additionally, human evaluations and factuality analyses confirm that \method's outputs are as helpful and often more factually grounded than those from comparable LLMs aligned with reward models. \method's success is enabled by stronger base language models, high-quality synthetic reference outputs, and better training algorithms, which collectively address the pitfalls that undermined earlier BLEU-based RL efforts. Overall, \method\ is a lightweight, cost-effective alternative to reward model-guided alignment, and we hope it facilitates future work in reward design and alignment strategies without expensive human preference supervision.
\section{Limitations} \label{sec:limitations}

While \method\ demonstrates promising results as a lightweight alternative to reward model-based alignment, our study has several limitations.
First, our experiments are limited in scope, covering only two model scales, a moderate-scale data pool (50K examples), and one string-overlap metric (BLEU). We do not fully explore the effects of scaling model size, data volume, training time, or training with other alternative metrics.
Second, we do not perform extensive hyperparameter tuning for each model and setup due to computational constraints.
Third, BLEU's reliance on surface-form $n$-gram overlap makes it sensitive to reference quality and vulnerable in domains with high lexical variation, such as mathematical reasoning and multilingual tasks.

\section{Acknowledgements}

We thank the University of Maryland Computational Linguistics and Information Processing
(CLIP) Lab for their feedback and support. We are grateful to Lambda AI for providing GPU resources for the training experiments. This project was partially supported by awards IIS2046248 and IIS-2312949 from the National Science Foundation (NSF) as well as an award from Open Philanthropy.

\bibliographystyle{plainnat}  
\bibliography{references}

\begin{thebibliography}{80}
\providecommand{\natexlab}[1]{#1}
\providecommand{\url}[1]{\texttt{#1}}
\expandafter\ifx\csname urlstyle\endcsname\relax
  \providecommand{\doi}[1]{doi: #1}\else
  \providecommand{\doi}{doi: \begingroup \urlstyle{rm}\Url}\fi

\bibitem[AI@Meta(2024)]{llama3modelcard}
AI@Meta.
\newblock Llama 3 model card.
\newblock 2024.
\newblock URL \url{https://github.com/meta-llama/llama3/blob/main/MODEL_CARD.md}.

\bibitem[Anthropic(2025)]{claude3_7-blog}
Anthropic.
\newblock Claude 3.7 sonnet and claude code.
\newblock \url{https://www.anthropic.com/news/claude-3-7-sonnet}, 2025.
\newblock Accessed: 2025-02-24.

\bibitem[Bahdanau et~al.(2017)Bahdanau, Brakel, Xu, Goyal, Lowe, Pineau, Courville, and Bengio]{bahdanau2017an}
Dzmitry Bahdanau, Philemon Brakel, Kelvin Xu, Anirudh Goyal, Ryan Lowe, Joelle Pineau, Aaron Courville, and Yoshua Bengio.
\newblock An actor-critic algorithm for sequence prediction.
\newblock In \emph{International Conference on Learning Representations}, 2017.
\newblock URL \url{https://openreview.net/forum?id=SJDaqqveg}.

\bibitem[Bai et~al.(2022)Bai, Jones, Ndousse, Askell, Chen, DasSarma, Drain, Fort, Ganguli, Henighan, Joseph, Kadavath, Kernion, Conerly, El-Showk, Elhage, Hatfield-Dodds, Hernandez, Hume, Johnston, Kravec, Lovitt, Nanda, Olsson, Amodei, Brown, Clark, McCandlish, Olah, Mann, and Kaplan]{bai2022traininghelpfulharmlessassistant}
Yuntao Bai, Andy Jones, Kamal Ndousse, Amanda Askell, Anna Chen, Nova DasSarma, Dawn Drain, Stanislav Fort, Deep Ganguli, Tom Henighan, Nicholas Joseph, Saurav Kadavath, Jackson Kernion, Tom Conerly, Sheer El-Showk, Nelson Elhage, Zac Hatfield-Dodds, Danny Hernandez, Tristan Hume, Scott Johnston, Shauna Kravec, Liane Lovitt, Neel Nanda, Catherine Olsson, Dario Amodei, Tom Brown, Jack Clark, Sam McCandlish, Chris Olah, Ben Mann, and Jared Kaplan.
\newblock Training a helpful and harmless assistant with reinforcement learning from human feedback, 2022.
\newblock URL \url{https://arxiv.org/abs/2204.05862}.

\bibitem[Callison-Burch et~al.(2006)Callison-Burch, Osborne, and Koehn]{callison-burch-etal-2006-evaluating}
Chris Callison-Burch, Miles Osborne, and Philipp Koehn.
\newblock Re-evaluating the role of {B}leu in machine translation research.
\newblock In Diana McCarthy and Shuly Wintner, editors, \emph{11th Conference of the {E}uropean Chapter of the Association for Computational Linguistics}, pages 249--256, Trento, Italy, April 2006. Association for Computational Linguistics.
\newblock URL \url{https://aclanthology.org/E06-1032/}.

\bibitem[Chen et~al.(2024)Chen, Chen, Liu, Jiang, and Wang]{Chen2024HumansOL}
Guiming~Hardy Chen, Shunian Chen, Ziche Liu, Feng Jiang, and Benyou Wang.
\newblock Humans or llms as the judge? a study on judgement bias.
\newblock In \emph{Conference on Empirical Methods in Natural Language Processing}, 2024.
\newblock URL \url{https://api.semanticscholar.org/CorpusID:267740522}.

\bibitem[Chen et~al.(2025)Chen, Tu, Wang, Liu, Tang, Du, Zhou, and Xie]{chen2025sftrlearlyinvestigation}
Hardy Chen, Haoqin Tu, Fali Wang, Hui Liu, Xianfeng Tang, Xinya Du, Yuyin Zhou, and Cihang Xie.
\newblock Sft or rl? an early investigation into training r1-like reasoning large vision-language models, 2025.
\newblock URL \url{https://arxiv.org/abs/2504.11468}.

\bibitem[DeepSeek-AI(2025{\natexlab{a}})]{deepseekai2025deepseekr1incentivizingreasoningcapability}
DeepSeek-AI.
\newblock Deepseek-r1: Incentivizing reasoning capability in llms via reinforcement learning, 2025{\natexlab{a}}.
\newblock URL \url{https://arxiv.org/abs/2501.12948}.

\bibitem[DeepSeek-AI(2025{\natexlab{b}})]{deepseekai2025deepseekv3technicalreport}
DeepSeek-AI.
\newblock Deepseek-v3 technical report, 2025{\natexlab{b}}.
\newblock URL \url{https://arxiv.org/abs/2412.19437}.

\bibitem[Evtikhiev et~al.(2023{\natexlab{a}})Evtikhiev, Bogomolov, Sokolov, and Bryksin]{Evtikhiev_2023_outbleu}
Mikhail Evtikhiev, Egor Bogomolov, Yaroslav Sokolov, and Timofey Bryksin.
\newblock Out of the bleu: How should we assess quality of the code generation models?
\newblock \emph{Journal of Systems and Software}, 203:\penalty0 111741, September 2023{\natexlab{a}}.
\newblock ISSN 0164-1212.
\newblock \doi{10.1016/j.jss.2023.111741}.
\newblock URL \url{http://dx.doi.org/10.1016/j.jss.2023.111741}.

\bibitem[Evtikhiev et~al.(2023{\natexlab{b}})Evtikhiev, Bogomolov, Sokolov, and Bryksin]{bleu-for-code}
Mikhail Evtikhiev, Egor Bogomolov, Yaroslav Sokolov, and Timofey Bryksin.
\newblock Out of the bleu: How should we assess quality of the code generation models?
\newblock \emph{J. Syst. Softw.}, 203\penalty0 (C), September 2023{\natexlab{b}}.
\newblock ISSN 0164-1212.
\newblock \doi{10.1016/j.jss.2023.111741}.
\newblock URL \url{https://doi.org/10.1016/j.jss.2023.111741}.

\bibitem[Freitag et~al.(2022)Freitag, Rei, Mathur, Lo, Stewart, Avramidis, Kocmi, Foster, Lavie, and Martins]{freitag-etal-2022-results}
Markus Freitag, Ricardo Rei, Nitika Mathur, Chi-kiu Lo, Craig Stewart, Eleftherios Avramidis, Tom Kocmi, George Foster, Alon Lavie, and Andr{\'e} F.~T. Martins.
\newblock Results of {WMT}22 metrics shared task: Stop using {BLEU} {--} neural metrics are better and more robust.
\newblock In Philipp Koehn, Lo{\"i}c Barrault, Ond{\v{r}}ej Bojar, Fethi Bougares, Rajen Chatterjee, Marta~R. Costa-juss{\`a}, Christian Federmann, Mark Fishel, Alexander Fraser, Markus Freitag, Yvette Graham, Roman Grundkiewicz, Paco Guzman, Barry Haddow, Matthias Huck, Antonio Jimeno~Yepes, Tom Kocmi, Andr{\'e} Martins, Makoto Morishita, Christof Monz, Masaaki Nagata, Toshiaki Nakazawa, Matteo Negri, Aur{\'e}lie N{\'e}v{\'e}ol, Mariana Neves, Martin Popel, Marco Turchi, and Marcos Zampieri, editors, \emph{Proceedings of the Seventh Conference on Machine Translation (WMT)}, pages 46--68, Abu Dhabi, United Arab Emirates (Hybrid), December 2022. Association for Computational Linguistics.
\newblock URL \url{https://aclanthology.org/2022.wmt-1.2/}.

\bibitem[Gemini-Team(2025)]{Gemini2_5-blog}
Gemini-Team.
\newblock Gemini 2.5: Our most intelligent ai model.
\newblock \url{https://blog.google/technology/google-deepmind/gemini-model-thinking-updates-march-2025/}, 2025.
\newblock Accessed: 2025-03-25.

\bibitem[Gurung and Lapata(2025)]{gurung2025learningreasonlongformstory}
Alexander Gurung and Mirella Lapata.
\newblock Learning to reason for long-form story generation, 2025.
\newblock URL \url{https://arxiv.org/abs/2503.22828}.

\bibitem[Haldar and Hockenmaier(2024)]{Haldar2024code}
Rajarshi Haldar and Julia Hockenmaier.
\newblock Analyzing the performance of large language models on code summarization.
\newblock In Nicoletta Calzolari, Min-Yen Kan, Veronique Hoste, Alessandro Lenci, Sakriani Sakti, and Nianwen Xue, editors, \emph{2024 Joint International Conference on Computational Linguistics, Language Resources and Evaluation, LREC-COLING 2024 - Main Conference Proceedings}, 2024 Joint International Conference on Computational Linguistics, Language Resources and Evaluation, LREC-COLING 2024 - Main Conference Proceedings, pages 995--1008. European Language Resources Association (ELRA), 2024.
\newblock This work is supported by Agriculture and Food Research Initiative (AFRI) grant no. 2020-67021-32799/project accession no.1024178 from the USDA National Institute of Food and Agriculture.; Joint 30th International Conference on Computational Linguistics and 14th International Conference on Language Resources and Evaluation, LREC-COLING 2024 ; Conference date: 20-05-2024 Through 25-05-2024.

\bibitem[Ive et~al.(2021)Ive, Wang, Fomicheva, and Specia]{ive-etal-2021-exploring}
Julia Ive, Zixu Wang, Marina Fomicheva, and Lucia Specia.
\newblock Exploring supervised and unsupervised rewards in machine translation.
\newblock In Paola Merlo, Jorg Tiedemann, and Reut Tsarfaty, editors, \emph{Proceedings of the 16th Conference of the European Chapter of the Association for Computational Linguistics: Main Volume}, pages 1908--1920, Online, April 2021. Association for Computational Linguistics.
\newblock \doi{10.18653/v1/2021.eacl-main.164}.
\newblock URL \url{https://aclanthology.org/2021.eacl-main.164/}.

\bibitem[Jiao et~al.(2024)Jiao, Qin, Liu, Chen, and Joty]{jiao-etal-2024-learning}
Fangkai Jiao, Chengwei Qin, Zhengyuan Liu, Nancy~F. Chen, and Shafiq Joty.
\newblock Learning planning-based reasoning by trajectories collection and process reward synthesizing.
\newblock In Yaser Al-Onaizan, Mohit Bansal, and Yun-Nung Chen, editors, \emph{Proceedings of the 2024 Conference on Empirical Methods in Natural Language Processing}, pages 334--350, Miami, Florida, USA, November 2024. Association for Computational Linguistics.
\newblock \doi{10.18653/v1/2024.emnlp-main.20}.
\newblock URL \url{https://aclanthology.org/2024.emnlp-main.20/}.

\bibitem[Kanerva et~al.(2019)Kanerva, R{\"o}nnqvist, Kekki, Salakoski, and Ginter]{kanerva-etal-2019-template}
Jenna Kanerva, Samuel R{\"o}nnqvist, Riina Kekki, Tapio Salakoski, and Filip Ginter.
\newblock Template-free data-to-text generation of {F}innish sports news.
\newblock In Mareike Hartmann and Barbara Plank, editors, \emph{Proceedings of the 22nd Nordic Conference on Computational Linguistics}, pages 242--252, Turku, Finland, September–October 2019. Link{\"o}ping University Electronic Press.
\newblock URL \url{https://aclanthology.org/W19-6125/}.

\bibitem[Khalifa et~al.(2021)Khalifa, Elsahar, and Dymetman]{khalifa2021a}
Muhammad Khalifa, Hady Elsahar, and Marc Dymetman.
\newblock A distributional approach to controlled text generation.
\newblock In \emph{International Conference on Learning Representations}, 2021.
\newblock URL \url{https://openreview.net/forum?id=jWkw45-9AbL}.

\bibitem[Kiegeland and Kreutzer(2021)]{kiegeland-kreutzer-2021-revisiting}
Samuel Kiegeland and Julia Kreutzer.
\newblock Revisiting the weaknesses of reinforcement learning for neural machine translation.
\newblock In Kristina Toutanova, Anna Rumshisky, Luke Zettlemoyer, Dilek Hakkani-Tur, Iz~Beltagy, Steven Bethard, Ryan Cotterell, Tanmoy Chakraborty, and Yichao Zhou, editors, \emph{Proceedings of the 2021 Conference of the North American Chapter of the Association for Computational Linguistics: Human Language Technologies}, pages 1673--1681, Online, June 2021. Association for Computational Linguistics.
\newblock \doi{10.18653/v1/2021.naacl-main.133}.
\newblock URL \url{https://aclanthology.org/2021.naacl-main.133/}.

\bibitem[Korbak et~al.(2022)Korbak, Elsahar, Kruszewski, and Dymetman]{korbak2022on}
Tomasz Korbak, Hady Elsahar, Germ{\'a}n Kruszewski, and Marc Dymetman.
\newblock On reinforcement learning and distribution matching for fine-tuning language models with no catastrophic forgetting.
\newblock In Alice~H. Oh, Alekh Agarwal, Danielle Belgrave, and Kyunghyun Cho, editors, \emph{Advances in Neural Information Processing Systems}, 2022.
\newblock URL \url{https://openreview.net/forum?id=XvI6h-s4un}.

\bibitem[Kryscinski et~al.(2019)Kryscinski, Keskar, McCann, Xiong, and Socher]{kryscinski-etal-2019-neural}
Wojciech Kryscinski, Nitish~Shirish Keskar, Bryan McCann, Caiming Xiong, and Richard Socher.
\newblock Neural text summarization: A critical evaluation.
\newblock In Kentaro Inui, Jing Jiang, Vincent Ng, and Xiaojun Wan, editors, \emph{Proceedings of the 2019 Conference on Empirical Methods in Natural Language Processing and the 9th International Joint Conference on Natural Language Processing (EMNLP-IJCNLP)}, pages 540--551, Hong Kong, China, November 2019. Association for Computational Linguistics.
\newblock \doi{10.18653/v1/D19-1051}.
\newblock URL \url{https://aclanthology.org/D19-1051/}.

\bibitem[Kwon et~al.(2023)Kwon, Li, Zhuang, Sheng, Zheng, Yu, Gonzalez, Zhang, and Stoica]{kwon2023efficient}
Woosuk Kwon, Zhuohan Li, Siyuan Zhuang, Ying Sheng, Lianmin Zheng, Cody~Hao Yu, Joseph~E. Gonzalez, Hao Zhang, and Ion Stoica.
\newblock Efficient memory management for large language model serving with pagedattention.
\newblock In \emph{Proceedings of the ACM SIGOPS 29th Symposium on Operating Systems Principles}, 2023.

\bibitem[Lambert et~al.(2024{\natexlab{a}})Lambert, Morrison, Pyatkin, Huang, Ivison, Brahman, Miranda, Liu, Dziri, Lyu, Gu, Malik, Graf, Hwang, Yang, Bras, Tafjord, Wilhelm, Soldaini, Smith, Wang, Dasigi, and Hajishirzi]{lambert2024tulu3}
Nathan Lambert, Jacob Morrison, Valentina Pyatkin, Shengyi Huang, Hamish Ivison, Faeze Brahman, Lester James~V. Miranda, Alisa Liu, Nouha Dziri, Shane Lyu, Yuling Gu, Saumya Malik, Victoria Graf, Jena~D. Hwang, Jiangjiang Yang, Ronan~Le Bras, Oyvind Tafjord, Chris Wilhelm, Luca Soldaini, Noah~A. Smith, Yizhong Wang, Pradeep Dasigi, and Hannaneh Hajishirzi.
\newblock Tülu 3: Pushing frontiers in open language model post-training.
\newblock 2024{\natexlab{a}}.

\bibitem[Lambert et~al.(2024{\natexlab{b}})Lambert, Pyatkin, Morrison, Miranda, Lin, Chandu, Dziri, Kumar, Zick, Choi, Smith, and Hajishirzi]{RewardBench}
Nathan Lambert, Valentina Pyatkin, Jacob Morrison, LJ~Miranda, Bill~Yuchen Lin, Khyathi Chandu, Nouha Dziri, Sachin Kumar, Tom Zick, Yejin Choi, Noah~A. Smith, and Hannaneh Hajishirzi.
\newblock Rewardbench: Evaluating reward models for language modeling.
\newblock \url{https://huggingface.co/spaces/allenai/reward-bench}, 2024{\natexlab{b}}.

\bibitem[Li et~al.(2024)Li, Chiang, Frick, Dunlap, Wu, Zhu, Gonzalez, and Stoica]{li2024crowdsourced}
Tianle Li, Wei-Lin Chiang, Evan Frick, Lisa Dunlap, Tianhao Wu, Banghua Zhu, Joseph~E Gonzalez, and Ion Stoica.
\newblock From crowdsourced data to high-quality benchmarks: Arena-hard and benchbuilder pipeline.
\newblock \emph{arXiv preprint arXiv:2406.11939}, 2024.

\bibitem[Li and Eisner(2009)]{li-eisner-2009-first}
Zhifei Li and Jason Eisner.
\newblock First- and second-order expectation semirings with applications to minimum-risk training on translation forests.
\newblock In Philipp Koehn and Rada Mihalcea, editors, \emph{Proceedings of the 2009 Conference on Empirical Methods in Natural Language Processing}, pages 40--51, Singapore, August 2009. Association for Computational Linguistics.
\newblock URL \url{https://aclanthology.org/D09-1005/}.

\bibitem[Lin et~al.(2025)Lin, Deng, Chandu, Ravichander, Pyatkin, Dziri, Bras, and Choi]{lin2025wildbench}
Bill~Yuchen Lin, Yuntian Deng, Khyathi Chandu, Abhilasha Ravichander, Valentina Pyatkin, Nouha Dziri, Ronan~Le Bras, and Yejin Choi.
\newblock Wildbench: Benchmarking {LLM}s with challenging tasks from real users in the wild.
\newblock In \emph{The Thirteenth International Conference on Learning Representations}, 2025.
\newblock URL \url{https://openreview.net/forum?id=MKEHCx25xp}.

\bibitem[Lin(2004)]{lin-2004-rouge}
Chin-Yew Lin.
\newblock {ROUGE}: A package for automatic evaluation of summaries.
\newblock In \emph{Text Summarization Branches Out}, pages 74--81, Barcelona, Spain, July 2004. Association for Computational Linguistics.
\newblock URL \url{https://aclanthology.org/W04-1013/}.

\bibitem[Liu et~al.(2016)Liu, Lowe, Serban, Noseworthy, Charlin, and Pineau]{liu-etal-2016-evaluate}
Chia-Wei Liu, Ryan Lowe, Iulian Serban, Mike Noseworthy, Laurent Charlin, and Joelle Pineau.
\newblock How {NOT} to evaluate your dialogue system: An empirical study of unsupervised evaluation metrics for dialogue response generation.
\newblock In Jian Su, Kevin Duh, and Xavier Carreras, editors, \emph{Proceedings of the 2016 Conference on Empirical Methods in Natural Language Processing}, pages 2122--2132, Austin, Texas, November 2016. Association for Computational Linguistics.
\newblock \doi{10.18653/v1/D16-1230}.
\newblock URL \url{https://aclanthology.org/D16-1230/}.

\bibitem[Liu et~al.(2025{\natexlab{a}})Liu, Chen, Li, Qi, Pang, Du, Lee, and Lin]{liu2025understandingr1zeroliketrainingcritical}
Zichen Liu, Changyu Chen, Wenjun Li, Penghui Qi, Tianyu Pang, Chao Du, Wee~Sun Lee, and Min Lin.
\newblock Understanding r1-zero-like training: A critical perspective, 2025{\natexlab{a}}.
\newblock URL \url{https://arxiv.org/abs/2503.20783}.

\bibitem[Liu et~al.(2025{\natexlab{b}})Liu, Sun, Zang, Dong, Cao, Duan, Lin, and Wang]{liu2025visualrftvisualreinforcementfinetuning}
Ziyu Liu, Zeyi Sun, Yuhang Zang, Xiaoyi Dong, Yuhang Cao, Haodong Duan, Dahua Lin, and Jiaqi Wang.
\newblock Visual-rft: Visual reinforcement fine-tuning, 2025{\natexlab{b}}.
\newblock URL \url{https://arxiv.org/abs/2503.01785}.

\bibitem[Mathur et~al.(2020)Mathur, Baldwin, and Cohn]{mathur-etal-2020-tangled}
Nitika Mathur, Timothy Baldwin, and Trevor Cohn.
\newblock Tangled up in {BLEU}: Reevaluating the evaluation of automatic machine translation evaluation metrics.
\newblock In Dan Jurafsky, Joyce Chai, Natalie Schluter, and Joel Tetreault, editors, \emph{Proceedings of the 58th Annual Meeting of the Association for Computational Linguistics}, pages 4984--4997, Online, July 2020. Association for Computational Linguistics.
\newblock \doi{10.18653/v1/2020.acl-main.448}.
\newblock URL \url{https://aclanthology.org/2020.acl-main.448/}.

\bibitem[Mostafazadeh et~al.(2020)Mostafazadeh, Kalyanpur, Moon, Buchanan, Berkowitz, Biran, and Chu-Carroll]{mostafazadeh-etal-2020-glucose}
Nasrin Mostafazadeh, Aditya Kalyanpur, Lori Moon, David Buchanan, Lauren Berkowitz, Or~Biran, and Jennifer Chu-Carroll.
\newblock {GLUCOSE}: {G}enera{L}ized and {CO}ntextualized story explanations.
\newblock In Bonnie Webber, Trevor Cohn, Yulan He, and Yang Liu, editors, \emph{Proceedings of the 2020 Conference on Empirical Methods in Natural Language Processing (EMNLP)}, pages 4569--4586, Online, November 2020. Association for Computational Linguistics.
\newblock \doi{10.18653/v1/2020.emnlp-main.370}.
\newblock URL \url{https://aclanthology.org/2020.emnlp-main.370/}.

\bibitem[Och(2003)]{och-2003-minimum}
Franz~Josef Och.
\newblock Minimum error rate training in statistical machine translation.
\newblock In \emph{Proceedings of the 41st Annual Meeting of the Association for Computational Linguistics}, pages 160--167, Sapporo, Japan, July 2003. Association for Computational Linguistics.
\newblock \doi{10.3115/1075096.1075117}.
\newblock URL \url{https://aclanthology.org/P03-1021/}.

\bibitem[OLMo et~al.(2024)OLMo, Walsh, Soldaini, Groeneveld, Lo, Arora, Bhagia, Gu, Huang, Jordan, Lambert, Schwenk, Tafjord, Anderson, Atkinson, Brahman, Clark, Dasigi, Dziri, Guerquin, Ivison, Koh, Liu, Malik, Merrill, Miranda, Morrison, Murray, Nam, Pyatkin, Rangapur, Schmitz, Skjonsberg, Wadden, Wilhelm, Wilson, Zettlemoyer, Farhadi, Smith, and Hajishirzi]{olmo20242olmo2furious}
Team OLMo, Pete Walsh, Luca Soldaini, Dirk Groeneveld, Kyle Lo, Shane Arora, Akshita Bhagia, Yuling Gu, Shengyi Huang, Matt Jordan, Nathan Lambert, Dustin Schwenk, Oyvind Tafjord, Taira Anderson, David Atkinson, Faeze Brahman, Christopher Clark, Pradeep Dasigi, Nouha Dziri, Michal Guerquin, Hamish Ivison, Pang~Wei Koh, Jiacheng Liu, Saumya Malik, William Merrill, Lester James~V. Miranda, Jacob Morrison, Tyler Murray, Crystal Nam, Valentina Pyatkin, Aman Rangapur, Michael Schmitz, Sam Skjonsberg, David Wadden, Christopher Wilhelm, Michael Wilson, Luke Zettlemoyer, Ali Farhadi, Noah~A. Smith, and Hannaneh Hajishirzi.
\newblock 2 olmo 2 furious.
\newblock 2024.
\newblock URL \url{https://arxiv.org/abs/2501.00656}.

\bibitem[OpenAI(2024)]{openai2024gpt4ocard}
OpenAI.
\newblock Gpt-4o system card, 2024.
\newblock URL \url{https://arxiv.org/abs/2410.21276}.

\bibitem[OpenAI(2025)]{openai2025o3o4}
OpenAI.
\newblock Openai o3 and o4-mini system card, 2025.
\newblock URL \url{https://cdn.openai.com/pdf/2221c875-02dc-4789-800b-e7758f3722c1/o3-and-o4-mini-system-card.pdf}.
\newblock Accessed: 2025-04-16.

\bibitem[Ouyang et~al.(2022{\natexlab{a}})Ouyang, Wu, Jiang, Almeida, Wainwright, Mishkin, Zhang, Agarwal, Slama, Ray, Schulman, Hilton, Kelton, Miller, Simens, Askell, Welinder, Christiano, Leike, and Lowe]{Ouyang2022TrainingLM}
Long Ouyang, Jeff Wu, Xu~Jiang, Diogo Almeida, Carroll~L. Wainwright, Pamela Mishkin, Chong Zhang, Sandhini Agarwal, Katarina Slama, Alex Ray, John Schulman, Jacob Hilton, Fraser Kelton, Luke~E. Miller, Maddie Simens, Amanda Askell, Peter Welinder, Paul~Francis Christiano, Jan Leike, and Ryan~J. Lowe.
\newblock Training language models to follow instructions with human feedback.
\newblock \emph{ArXiv}, abs/2203.02155, 2022{\natexlab{a}}.
\newblock URL \url{https://api.semanticscholar.org/CorpusID:246426909}.

\bibitem[Ouyang et~al.(2022{\natexlab{b}})Ouyang, Wu, Jiang, Almeida, Wainwright, Mishkin, Zhang, Agarwal, Slama, Ray, et~al.]{ouyang2022training}
Long Ouyang, Jeffrey Wu, Xu~Jiang, Diogo Almeida, Carroll Wainwright, Pamela Mishkin, Chong Zhang, Sandhini Agarwal, Katarina Slama, Alex Ray, et~al.
\newblock Training language models to follow instructions with human feedback.
\newblock \emph{Advances in neural information processing systems}, 35:\penalty0 27730--27744, 2022{\natexlab{b}}.

\bibitem[Papineni et~al.(2002)Papineni, Roukos, Ward, and Zhu]{bleu}
Kishore Papineni, Salim Roukos, Todd Ward, and Wei-Jing Zhu.
\newblock Bleu: a method for automatic evaluation of machine translation.
\newblock In \emph{Proceedings of the 40th Annual Meeting on Association for Computational Linguistics}, ACL '02, page 311–318, USA, 2002. Association for Computational Linguistics.
\newblock \doi{10.3115/1073083.1073135}.
\newblock URL \url{https://doi.org/10.3115/1073083.1073135}.

\bibitem[Qwen-Team(2025)]{qwen2025qwen25technicalreport}
Qwen-Team.
\newblock Qwen2.5 technical report, 2025.
\newblock URL \url{https://arxiv.org/abs/2412.15115}.

\bibitem[Ranzato et~al.(2016)Ranzato, Chopra, Auli, and Zaremba]{ranzato2016sequenceleveltrainingrecurrent}
Marc'Aurelio Ranzato, Sumit Chopra, Michael Auli, and Wojciech Zaremba.
\newblock Sequence level training with recurrent neural networks.
\newblock 2016.
\newblock URL \url{https://arxiv.org/abs/1511.06732}.

\bibitem[Reiter(2018)]{reiter-2018-structured}
Ehud Reiter.
\newblock A structured review of the validity of {BLEU}.
\newblock \emph{Computational Linguistics}, 44\penalty0 (3):\penalty0 393--401, September 2018.
\newblock \doi{10.1162/coli_a_00322}.
\newblock URL \url{https://aclanthology.org/J18-3002/}.

\bibitem[Rennie et~al.(2017)Rennie, Marcheret, Mroueh, Ross, and Goel]{rennie2017}
Steven~J. Rennie, Etienne Marcheret, Youssef Mroueh, Jerret Ross, and Vaibhava Goel.
\newblock Self-critical sequence training for image captioning.
\newblock In \emph{2017 IEEE Conference on Computer Vision and Pattern Recognition (CVPR)}, pages 1179--1195, 2017.
\newblock \doi{10.1109/CVPR.2017.131}.

\bibitem[Saito et~al.(2023)Saito, Wachi, Wataoka, and Akimoto]{saito2023verbositybiaspreferencelabeling}
Keita Saito, Akifumi Wachi, Koki Wataoka, and Youhei Akimoto.
\newblock Verbosity bias in preference labeling by large language models, 2023.
\newblock URL \url{https://arxiv.org/abs/2310.10076}.

\bibitem[Schulman et~al.(2017)Schulman, Wolski, Dhariwal, Radford, and Klimov]{schulman2017proximalpolicyoptimizationalgorithms}
John Schulman, Filip Wolski, Prafulla Dhariwal, Alec Radford, and Oleg Klimov.
\newblock Proximal policy optimization algorithms, 2017.
\newblock URL \url{https://arxiv.org/abs/1707.06347}.

\bibitem[Shailya et~al.(2025)Shailya, Rajpal, Krishnan, and Ravindran]{shailya2025lextevaluatingtrustworthinessnatural}
Krithi Shailya, Shreya Rajpal, Gokul~S Krishnan, and Balaraman Ravindran.
\newblock Lext: Towards evaluating trustworthiness of natural language explanations, 2025.
\newblock URL \url{https://arxiv.org/abs/2504.06227}.

\bibitem[Shao et~al.(2025)Shao, Li, Xin, Geng, Wang, Oh, Du, Lambert, Min, Krishna, Tsvetkov, Hajishirzi, Koh, and Zettlemoyer]{shao2025spurious}
Rulin Shao, Shuyue~Stella Li, Rui Xin, Scott Geng, Yiping Wang, Sewoong Oh, Simon~Shaolei Du, Nathan Lambert, Sewon Min, Ranjay Krishna, Yulia Tsvetkov, Hannaneh Hajishirzi, Pang~Wei Koh, and Luke Zettlemoyer.
\newblock Spurious rewards: Rethinking training signals in rlvr.
\newblock \url{https://rethink-rlvr.notion.site/Spurious-Rewards-Rethinking-Training-Signals-in-RLVR-1f4df34dac1880948858f95aeb88872f}, 2025.
\newblock Notion Blog.

\bibitem[Shao et~al.(2024)Shao, Wang, Zhu, Xu, Song, Bi, Zhang, Zhang, Li, Wu, and Guo]{shao2024deepseekmathpushinglimitsmathematical}
Zhihong Shao, Peiyi Wang, Qihao Zhu, Runxin Xu, Junxiao Song, Xiao Bi, Haowei Zhang, Mingchuan Zhang, Y.~K. Li, Y.~Wu, and Daya Guo.
\newblock Deepseekmath: Pushing the limits of mathematical reasoning in open language models, 2024.
\newblock URL \url{https://arxiv.org/abs/2402.03300}.

\bibitem[Singhal et~al.(2024)Singhal, Goyal, Xu, and Durrett]{singhal2024longwaygoinvestigating}
Prasann Singhal, Tanya Goyal, Jiacheng Xu, and Greg Durrett.
\newblock A long way to go: Investigating length correlations in rlhf, 2024.
\newblock URL \url{https://arxiv.org/abs/2310.03716}.

\bibitem[Smith and Eisner(2006)]{smith-eisner-2006-minimum}
David~A. Smith and Jason Eisner.
\newblock Minimum risk annealing for training log-linear models.
\newblock In \emph{Proceedings of the {COLING}/{ACL} 2006 Main Conference Poster Sessions}, pages 787--794, Sydney, Australia, July 2006. Association for Computational Linguistics.
\newblock URL \url{https://aclanthology.org/P06-2101/}.

\bibitem[Song et~al.(2024)Song, Kim, and Iyyer]{song-etal-2024-veriscore}
Yixiao Song, Yekyung Kim, and Mohit Iyyer.
\newblock {V}eri{S}core: Evaluating the factuality of verifiable claims in long-form text generation.
\newblock In Yaser Al-Onaizan, Mohit Bansal, and Yun-Nung Chen, editors, \emph{Findings of the Association for Computational Linguistics: EMNLP 2024}, pages 9447--9474, Miami, Florida, USA, November 2024. Association for Computational Linguistics.
\newblock \doi{10.18653/v1/2024.findings-emnlp.552}.
\newblock URL \url{https://aclanthology.org/2024.findings-emnlp.552/}.

\bibitem[Sprague et~al.(2025)Sprague, Yin, Rodriguez, Jiang, Wadhwa, Singhal, Zhao, Ye, Mahowald, and Durrett]{sprague2025to}
Zayne~Rea Sprague, Fangcong Yin, Juan~Diego Rodriguez, Dongwei Jiang, Manya Wadhwa, Prasann Singhal, Xinyu Zhao, Xi~Ye, Kyle Mahowald, and Greg Durrett.
\newblock To cot or not to cot? chain-of-thought helps mainly on math and symbolic reasoning.
\newblock In \emph{The Thirteenth International Conference on Learning Representations}, 2025.
\newblock URL \url{https://openreview.net/forum?id=w6nlcS8Kkn}.

\bibitem[Stiennon et~al.(2022)Stiennon, Ouyang, Wu, Ziegler, Lowe, Voss, Radford, Amodei, and Christiano]{stiennon2022learningsummarizehumanfeedback}
Nisan Stiennon, Long Ouyang, Jeff Wu, Daniel~M. Ziegler, Ryan Lowe, Chelsea Voss, Alec Radford, Dario Amodei, and Paul Christiano.
\newblock Learning to summarize from human feedback, 2022.
\newblock URL \url{https://arxiv.org/abs/2009.01325}.

\bibitem[Team(2025)]{qwen3technical2025}
Qwen Team.
\newblock Qwen3 technical report.
\newblock \url{https://github.com/QwenLM/Qwen3/blob/main/Qwen3_Technical_Report.pdf}, 2025.

\bibitem[Teknium(2024)]{openhermes2024}
Teknium.
\newblock Openhermes 2.5 - mistral 7b.
\newblock \url{https://huggingface.co/teknium/OpenHermes-2.5-Mistral-7B}, 2024.

\bibitem[Thai et~al.(2022)Thai, Karpinska, Krishna, Ray, Inghilleri, Wieting, and Iyyer]{thai2022exploring}
Katherine Thai, Marzena Karpinska, Kalpesh Krishna, Bill Ray, Moira Inghilleri, John Wieting, and Mohit Iyyer.
\newblock Exploring document-level literary machine translation with parallel paragraphs from world literature.
\newblock \emph{Conference on Empirical Methods in Natural Language Processing}, 2022.

\bibitem[Thulke et~al.(2023)Thulke, Daheim, Dugast, and Ney]{Thulke2023GroundedDialog}
David Thulke, Nico Daheim, Christian Dugast, and Hermann Ney.
\newblock Task-oriented document-grounded dialog systems by hltpr@rwth for dstc9 and dstc10.
\newblock \emph{IEEE/ACM Trans. Audio, Speech and Lang. Proc.}, 32:\penalty0 733–741, April 2023.
\newblock ISSN 2329-9290.
\newblock \doi{10.1109/TASLP.2023.3267832}.
\newblock URL \url{https://doi.org/10.1109/TASLP.2023.3267832}.

\bibitem[Touvron et~al.(2023)Touvron, Martin, Stone, Albert, Almahairi, Babaei, lay Bashlykov, Batra, Bhargava, Bhosale, Bikel, Blecher, tian Cant{\'o}n~Ferrer, Chen, Cucurull, Esiobu, Fernandes, Fu, Fu, Fuller, Gao, Goswami, Goyal, Hartshorn, Hosseini, Hou, Inan, Kardas, Kerkez, Khabsa, Kloumann, Korenev, Koura, Lachaux, Lavril, Lee, Liskovich, Lu, Mao, Martinet, Mihaylov, Mishra, Molybog, Nie, Poulton, Reizenstein, Rungta, Saladi, Schelten, Silva, Smith, Subramanian, Tan, Tang, Taylor, Williams, Kuan, Xu, Yan, Zarov, Zhang, Fan, Kambadur, Narang, Rodriguez, Stojnic, Edunov, and Scialom]{Touvron2023Llama2O}
Hugo Touvron, Louis Martin, Kevin~R. Stone, Peter Albert, Amjad Almahairi, Yasmine Babaei, Niko lay Bashlykov, Soumya Batra, Prajjwal Bhargava, Shruti Bhosale, Daniel~M. Bikel, Lukas Blecher, Cris tian Cant{\'o}n~Ferrer, Moya Chen, Guillem Cucurull, David Esiobu, Jude Fernandes, Jeremy Fu, Wenyin Fu, Brian Fuller, Cynthia Gao, Vedanuj Goswami, Naman Goyal, Anthony~S. Hartshorn, Saghar Hosseini, Rui Hou, Hakan Inan, Marcin Kardas, Viktor Kerkez, Madian Khabsa, Isabel~M. Kloumann, A.~V. Korenev, Punit~Singh Koura, Marie-Anne Lachaux, Thibaut Lavril, Jenya Lee, Diana Liskovich, Yinghai Lu, Yuning Mao, Xavier Martinet, Todor Mihaylov, Pushkar Mishra, Igor Molybog, Yixin Nie, Andrew Poulton, Jeremy Reizenstein, Rashi Rungta, Kalyan Saladi, Alan Schelten, Ruan Silva, Eric~Michael Smith, R.~Subramanian, Xia Tan, Binh Tang, Ross Taylor, Adina Williams, Jian~Xiang Kuan, Puxin Xu, Zhengxu Yan, Iliyan Zarov, Yuchen Zhang, Angela Fan, Melissa Hall~Melanie Kambadur, Sharan Narang, Aur'elien Rodriguez, Robert Stojnic,
  Sergey Edunov, and Thomas Scialom.
\newblock Llama 2: Open foundation and fine-tuned chat models.
\newblock \emph{ArXiv}, abs/2307.09288, 2023.
\newblock URL \url{https://api.semanticscholar.org/CorpusID:259950998}.

\bibitem[Wang et~al.(2020)Wang, Cho, and Lewis]{wang2020askingansweringquestionsevaluate}
Alex Wang, Kyunghyun Cho, and Mike Lewis.
\newblock Asking and answering questions to evaluate the factual consistency of summaries, 2020.
\newblock URL \url{https://arxiv.org/abs/2004.04228}.

\bibitem[Wang et~al.(2025{\natexlab{a}})Wang, Yang, Zeng, Ren, Liu, Peng, Cheng, He, Wang, Gao, Chen, Wang, Du, and Shen]{wang2025reinforcementlearningreasoninglarge}
Yiping Wang, Qing Yang, Zhiyuan Zeng, Liliang Ren, Lucas Liu, Baolin Peng, Hao Cheng, Xuehai He, Kuan Wang, Jianfeng Gao, Weizhu Chen, Shuohang Wang, Simon~Shaolei Du, and Yelong Shen.
\newblock Reinforcement learning for reasoning in large language models with one training example, 2025{\natexlab{a}}.
\newblock URL \url{https://arxiv.org/abs/2504.20571}.

\bibitem[Wang et~al.(2025{\natexlab{b}})Wang, Wang, Wang, Zhang, Li, Yang, Yu, Nguyen, Liu, Gottlieb, Lam, Lu, Cho, Wu, Fei-Fei, Wang, Choi, and Li]{wang2025ragenunderstandingselfevolutionllm}
Zihan Wang, Kangrui Wang, Qineng Wang, Pingyue Zhang, Linjie Li, Zhengyuan Yang, Kefan Yu, Minh~Nhat Nguyen, Licheng Liu, Eli Gottlieb, Monica Lam, Yiping Lu, Kyunghyun Cho, Jiajun Wu, Li~Fei-Fei, Lijuan Wang, Yejin Choi, and Manling Li.
\newblock Ragen: Understanding self-evolution in llm agents via multi-turn reinforcement learning, 2025{\natexlab{b}}.
\newblock URL \url{https://arxiv.org/abs/2504.20073}.

\bibitem[Watts et~al.(2024)Watts, Gumma, Yadavalli, Seshadri, Swaminathan, and Sitaram]{Watts2024PARIKSHAAL}
Ishaan Watts, Varun Gumma, Aditya Yadavalli, Vivek Seshadri, Manohar Swaminathan, and Sunayana Sitaram.
\newblock Pariksha: A large-scale investigation of human-llm evaluator agreement on multilingual and multi-cultural data.
\newblock In \emph{Conference on Empirical Methods in Natural Language Processing}, 2024.
\newblock URL \url{https://api.semanticscholar.org/CorpusID:270688323}.

\bibitem[Wei(2019)]{wei2019conductingbettervalidationstudies}
Johnny Tian-Zheng Wei.
\newblock On conducting better validation studies of automatic metrics in natural language generation evaluation, 2019.
\newblock URL \url{https://arxiv.org/abs/1907.13362}.

\bibitem[Wieting et~al.(2019)Wieting, Berg-Kirkpatrick, Gimpel, and Neubig]{wieting-etal-2019-beyond}
John Wieting, Taylor Berg-Kirkpatrick, Kevin Gimpel, and Graham Neubig.
\newblock Beyond {BLEU}: Training neural machine translation with semantic similarity.
\newblock In Anna Korhonen, David Traum, and Llu{\'i}s M{\`a}rquez, editors, \emph{Proceedings of the 57th Annual Meeting of the Association for Computational Linguistics}, pages 4344--4355, Florence, Italy, July 2019. Association for Computational Linguistics.
\newblock \doi{10.18653/v1/P19-1427}.
\newblock URL \url{https://aclanthology.org/P19-1427/}.

\bibitem[Williams(1992)]{reinforce-1992}
Ronald~J. Williams.
\newblock Simple statistical gradient-following algorithms for connectionist reinforcement learning.
\newblock \emph{Mach. Learn.}, 8\penalty0 (3–4):\penalty0 229–256, May 1992.
\newblock ISSN 0885-6125.
\newblock \doi{10.1007/BF00992696}.
\newblock URL \url{https://doi.org/10.1007/BF00992696}.

\bibitem[Wu et~al.(2018)Wu, Tian, Qin, Lai, and Liu]{Wu2018ASO}
Lijun Wu, Fei Tian, Tao Qin, Jianhuang Lai, and Tie-Yan Liu.
\newblock A study of reinforcement learning for neural machine translation.
\newblock In \emph{Conference on Empirical Methods in Natural Language Processing}, 2018.
\newblock URL \url{https://api.semanticscholar.org/CorpusID:52100616}.

\bibitem[Xu et~al.(2023{\natexlab{a}})Xu, Song, Iyyer, and Choi]{xu-etal-2023-critical}
Fangyuan Xu, Yixiao Song, Mohit Iyyer, and Eunsol Choi.
\newblock A critical evaluation of evaluations for long-form question answering.
\newblock In Anna Rogers, Jordan Boyd-Graber, and Naoaki Okazaki, editors, \emph{Proceedings of the 61st Annual Meeting of the Association for Computational Linguistics (Volume 1: Long Papers)}, pages 3225--3245, Toronto, Canada, July 2023{\natexlab{a}}. Association for Computational Linguistics.
\newblock \doi{10.18653/v1/2023.acl-long.181}.
\newblock URL \url{https://aclanthology.org/2023.acl-long.181/}.

\bibitem[Xu et~al.(2023{\natexlab{b}})Xu, Song, Iyyer, and Choi]{xu2023critical}
Fangyuan Xu, Yixiao Song, Mohit Iyyer, and Eunsol Choi.
\newblock A critical evaluation of evaluations for long-form question answering.
\newblock \emph{Association for Computational Linguistics}, 2023{\natexlab{b}}.

\bibitem[Xu et~al.(2025{\natexlab{a}})Xu, Peng, Awadalla, Chen, Chen, Gao, Kim, Li, Ren, Shen, Wang, Xu, Gao, and Chen]{xu2025phi4minireasoningexploringlimitssmall}
Haoran Xu, Baolin Peng, Hany Awadalla, Dongdong Chen, Yen-Chun Chen, Mei Gao, Young~Jin Kim, Yunsheng Li, Liliang Ren, Yelong Shen, Shuohang Wang, Weijian Xu, Jianfeng Gao, and Weizhu Chen.
\newblock Phi-4-mini-reasoning: Exploring the limits of small reasoning language models in math, 2025{\natexlab{a}}.
\newblock URL \url{https://arxiv.org/abs/2504.21233}.

\bibitem[Xu et~al.(2019)Xu, Wang, Tang, Duan, Yang, Zeng, Zhou, and Sun]{xu-etal-2019-asking}
Jingjing Xu, Yuechen Wang, Duyu Tang, Nan Duan, Pengcheng Yang, Qi~Zeng, Ming Zhou, and Xu~Sun.
\newblock Asking clarification questions in knowledge-based question answering.
\newblock In Kentaro Inui, Jing Jiang, Vincent Ng, and Xiaojun Wan, editors, \emph{Proceedings of the 2019 Conference on Empirical Methods in Natural Language Processing and the 9th International Joint Conference on Natural Language Processing (EMNLP-IJCNLP)}, pages 1618--1629, Hong Kong, China, November 2019. Association for Computational Linguistics.
\newblock \doi{10.18653/v1/D19-1172}.
\newblock URL \url{https://aclanthology.org/D19-1172/}.

\bibitem[Xu et~al.(2025{\natexlab{b}})Xu, Jiang, Niu, Deng, Poovendran, Choi, and Lin]{xu2025magpie}
Zhangchen Xu, Fengqing Jiang, Luyao Niu, Yuntian Deng, Radha Poovendran, Yejin Choi, and Bill~Yuchen Lin.
\newblock Magpie: Alignment data synthesis from scratch by prompting aligned {LLM}s with nothing.
\newblock In \emph{The Thirteenth International Conference on Learning Representations}, 2025{\natexlab{b}}.
\newblock URL \url{https://openreview.net/forum?id=Pnk7vMbznK}.

\bibitem[Yang et~al.(2024)Yang, Zhang, Hui, Gao, Yu, Li, Liu, Tu, Zhou, Lin, Lu, Xue, Lin, Liu, Ren, and Zhang]{yang2024qwen25mathtechnicalreportmathematical}
An~Yang, Beichen Zhang, Binyuan Hui, Bofei Gao, Bowen Yu, Chengpeng Li, Dayiheng Liu, Jianhong Tu, Jingren Zhou, Junyang Lin, Keming Lu, Mingfeng Xue, Runji Lin, Tianyu Liu, Xingzhang Ren, and Zhenru Zhang.
\newblock Qwen2.5-math technical report: Toward mathematical expert model via self-improvement, 2024.
\newblock URL \url{https://arxiv.org/abs/2409.12122}.

\bibitem[Yu et~al.(2025)Yu, Zhang, Zhu, Yuan, Zuo, Yue, Fan, Liu, Liu, Liu, Lin, Lin, Ma, Sheng, Tong, Zhang, Zhang, Zhang, Zhu, Zhu, Chen, Chen, Wang, Yu, Dai, Song, Wei, Zhou, Liu, Ma, Zhang, Yan, Qiao, Wu, and Wang]{yu2025dapoopensourcellmreinforcement}
Qiying Yu, Zheng Zhang, Ruofei Zhu, Yufeng Yuan, Xiaochen Zuo, Yu~Yue, Tiantian Fan, Gaohong Liu, Lingjun Liu, Xin Liu, Haibin Lin, Zhiqi Lin, Bole Ma, Guangming Sheng, Yuxuan Tong, Chi Zhang, Mofan Zhang, Wang Zhang, Hang Zhu, Jinhua Zhu, Jiaze Chen, Jiangjie Chen, Chengyi Wang, Hongli Yu, Weinan Dai, Yuxuan Song, Xiangpeng Wei, Hao Zhou, Jingjing Liu, Wei-Ying Ma, Ya-Qin Zhang, Lin Yan, Mu~Qiao, Yonghui Wu, and Mingxuan Wang.
\newblock Dapo: An open-source llm reinforcement learning system at scale, 2025.
\newblock URL \url{https://arxiv.org/abs/2503.14476}.

\bibitem[Zens et~al.(2007)Zens, Hasan, and Ney]{zens-etal-2007-systematic}
Richard Zens, Sa{\v{s}}a Hasan, and Hermann Ney.
\newblock A systematic comparison of training criteria for statistical machine translation.
\newblock In Jason Eisner, editor, \emph{Proceedings of the 2007 Joint Conference on Empirical Methods in Natural Language Processing and Computational Natural Language Learning ({EMNLP}-{C}o{NLL})}, pages 524--532, Prague, Czech Republic, June 2007. Association for Computational Linguistics.
\newblock URL \url{https://aclanthology.org/D07-1055/}.

\bibitem[Zhang et~al.(2025)Zhang, Liu, Qin, Naumann, and Poon]{zhang2025medrlvremergingmedicalreasoning}
Sheng Zhang, Qianchu Liu, Guanghui Qin, Tristan Naumann, and Hoifung Poon.
\newblock Med-rlvr: Emerging medical reasoning from a 3b base model via reinforcement learning, 2025.
\newblock URL \url{https://arxiv.org/abs/2502.19655}.

\bibitem[Zhang* et~al.(2020)Zhang*, Kishore*, Wu*, Weinberger, and Artzi]{Zhang*2020BERTScore:}
Tianyi Zhang*, Varsha Kishore*, Felix Wu*, Kilian~Q. Weinberger, and Yoav Artzi.
\newblock Bertscore: Evaluating text generation with bert.
\newblock In \emph{International Conference on Learning Representations}, 2020.
\newblock URL \url{https://openreview.net/forum?id=SkeHuCVFDr}.

\bibitem[Zheng et~al.(2023)Zheng, Chiang, Sheng, Zhuang, Wu, Zhuang, Lin, Li, Li, Xing, Zhang, Gonzalez, and Stoica]{zheng2023judging}
Lianmin Zheng, Wei-Lin Chiang, Ying Sheng, Siyuan Zhuang, Zhanghao Wu, Yonghao Zhuang, Zi~Lin, Zhuohan Li, Dacheng Li, Eric.~P Xing, Hao Zhang, Joseph~E. Gonzalez, and Ion Stoica.
\newblock Judging llm-as-a-judge with mt-bench and chatbot arena, 2023.

\bibitem[Zhou et~al.(2023)Zhou, Liu, Xu, Iyer, Sun, Mao, Ma, Efrat, Yu, YU, Zhang, Ghosh, Lewis, Zettlemoyer, and Levy]{zhou2023lima}
Chunting Zhou, Pengfei Liu, Puxin Xu, Srini Iyer, Jiao Sun, Yuning Mao, Xuezhe Ma, Avia Efrat, Ping Yu, LILI YU, Susan Zhang, Gargi Ghosh, Mike Lewis, Luke Zettlemoyer, and Omer Levy.
\newblock {LIMA}: Less is more for alignment.
\newblock In \emph{Thirty-seventh Conference on Neural Information Processing Systems}, 2023.
\newblock URL \url{https://openreview.net/forum?id=KBMOKmX2he}.

\end{thebibliography}

\appendix
\newpage

\section{Chatbot Arena human preference analysis} \label{appendix:pref-analysis}

\subsection{Obtaining the 900 subset}  \label{appendix-pref-analysis-getting-900}

To obtain the 900 subset we use for our human preference analysis, we first filter out all examples where human preference label is ``tie'', and/or the instruction or either of $O_X$ and $O_Y$ are shorter than 10 words or longer than 512 tokens. Then, we randomly sample 1,000 prompts and classify them using the approach described in \ref{sec:app-diff-domain}, then filter out 100 examples that receive a ``N/A'' label to reduce noise.

\subsection{Multi-reference configurations} \label{appendix:pref-analysis-multiref}

Below are the configurations we use for BLEU's multi-reference setups:
\begin{enumerate}
    \item 2 ref: Gemini-2.5-Pro, Deepseek-V3
    \item 3 ref: Gemini-2.5-Pro, Deepseek-V3, o4-mini
    \item 4 ref: Gemini-2.5-Pro, Deepseek-V3, o4-mini, Claude-3.7-Sonnet
    \item 5 ref: Gemini-2.5-Pro, Deepseek-V3, o4-mini, Claude-3.7-Sonnet, Qwen-Max
    \item 6 ref: Gemini-2.5-Pro, Deepseek-V3, o4-mini, Claude-3.7-Sonnet, Qwen-Max, GPT-4o
\end{enumerate}

\subsection{Ablation on components of BLEU} \label{appendix:pref-analysis-bleu-ablation}

We separately show the effect of n-gram precision and brevity penalty in \autoref{fig:agreement_category_all}. Neither component alone achieves high agreement with human labels.

\subsection{Impact of reference length on human agreement}  \label{sec:appendix-pref-analysis-len-impact}

\begin{figure}[htbp]
    \centering
    \includegraphics[width=\textwidth]{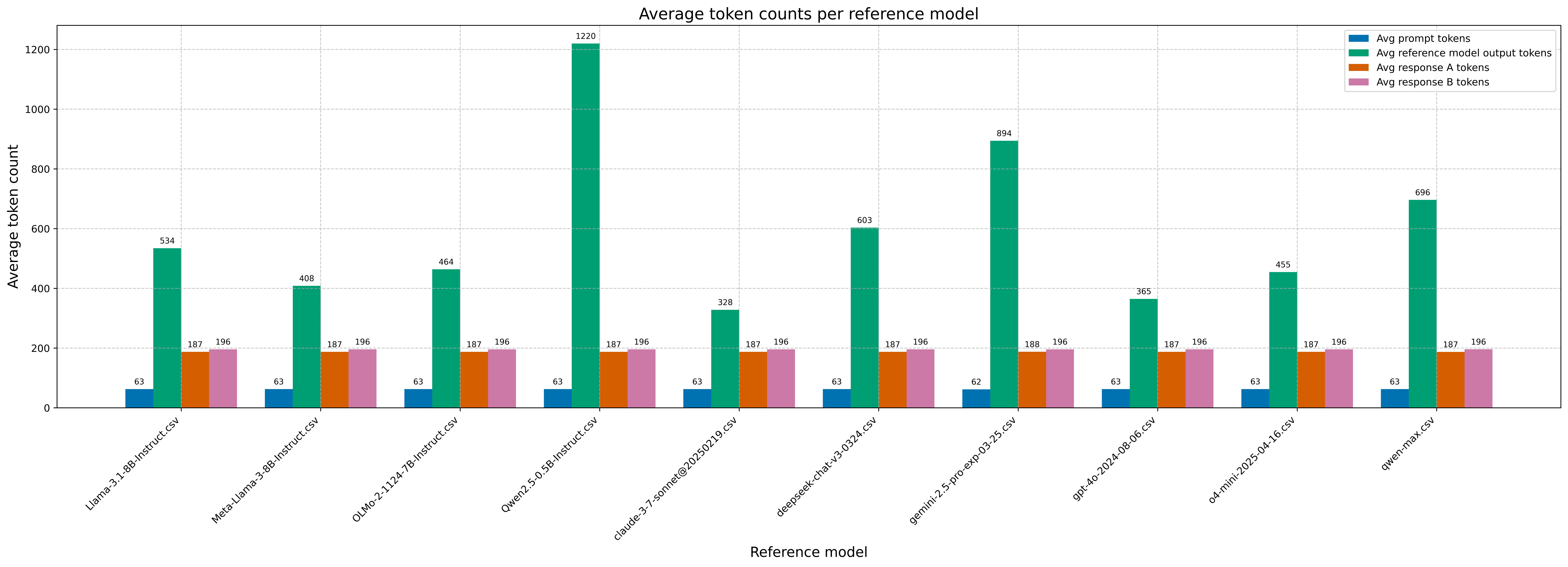}
    \caption{Token counts for difference reference model outputs, prompts, and the two model outputs to be scored.}
    \label{fig:ref-model-output-toks}
\end{figure}

In \autoref{fig:ref-model-output-toks}, we show the average number of tokens in the reference outputs generated by different LLMs, compared to the average tokens in $O_X$ and $O_Y$.

\subsection{Additional example for BLEU's $n$-gram attribution}  \label{sec:appendix-pref-analysis-ngram-attribution}

\begin{figure}{}
    \centering
    \includegraphics[width=0.9\textwidth]{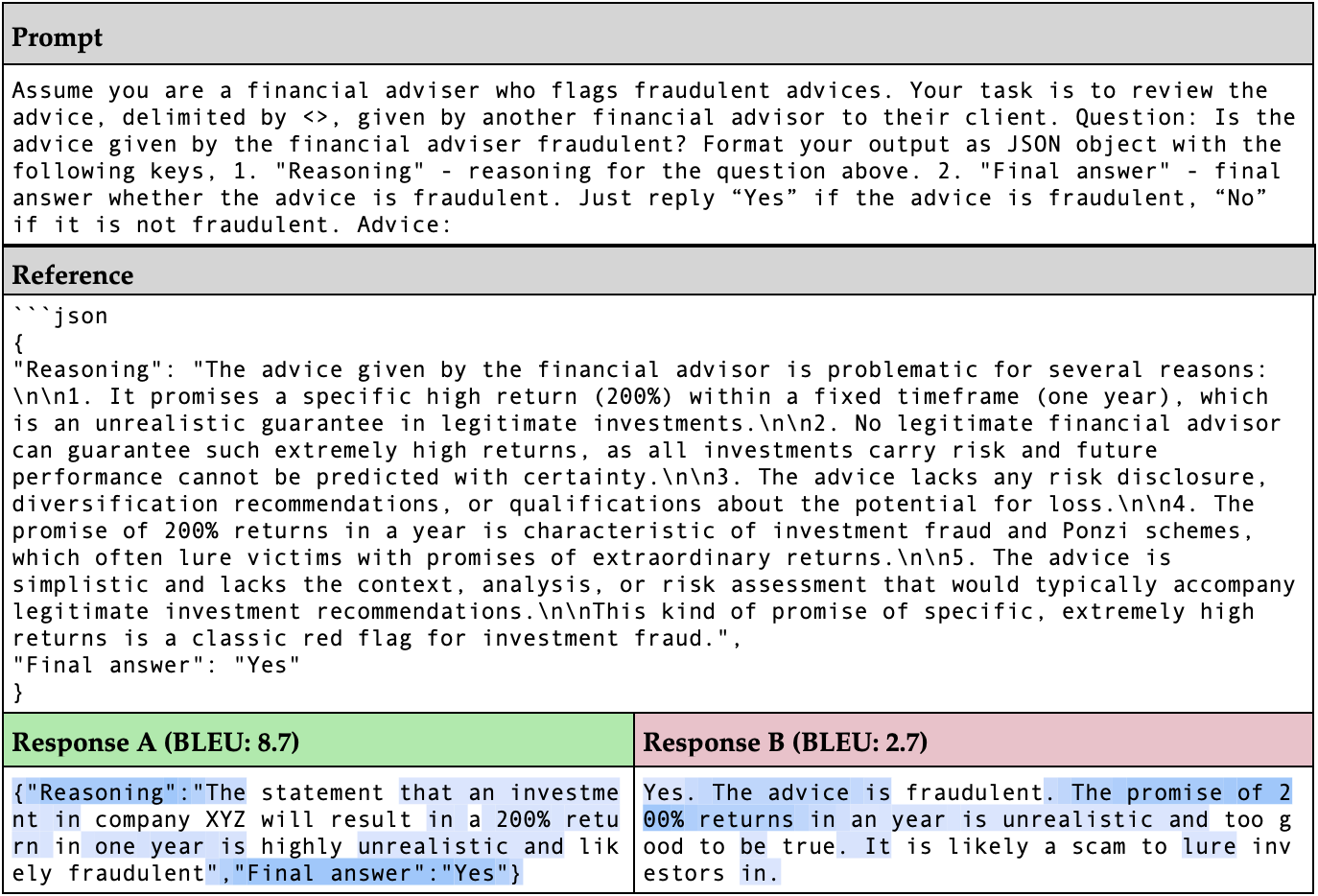}
    \caption{Additional example for BLEU's $n$-gram attribution on cases where it agrees with human preferences.}
    \label{fig:bleu-example-2}
\end{figure}

Please see \autoref{fig:bleu-example-2} for another example. Here, BLEU captures both formatting (e.g., JSON structure) and factual accuracy (i.e., providing the correct answer based on reasoning). This suggests that the BLEU score can serve as a useful signal for learning factual accuracy and format consistency.

\begin{table}[h!]
\centering
\footnotesize
\arrayrulecolor{black}
\caption{Hierarchy table of domains and their subcategories, grouped by characteristics. Each subcategory is sourced from \cite{lin2025wildbench}.}
\label{tab:data_domain_subcategory}
\begin{tabular}{ccc}
\toprule
\textbf{Domain} & \textbf{Subcategory} & \textbf{\# of examples} \\
\midrule
\multirow{4}{*}{QA} & Information seeking & 6516 \\
 & Advice seeking & 824 \\
 & Document QA & 411 \\
 & Data analysis & 189 \\
\midrule
\multirow{4}{*}{Writing} & Writing & 2565 \\
 & Roleplaying & 1112 \\
 & Summarization & 390 \\
 & Editing & 132 \\
\midrule
\multirow{3}{*}{Math and Reasoning} & Math & 1302 \\
 & Reasoning & 1064 \\
 & Classification & 662 \\
\midrule
\multirow{1}{*}{Code} & Coding and Debugging & 2095 \\
\midrule
\multirow{1}{*}{Multilingual} & Multilingual & 278 \\
\midrule
\multirow{2}{*}{Planning} & Planning & 257 \\
 & brainstorming & 399 \\
\bottomrule
\end{tabular}
\end{table}

\subsection{Human agreement rates across different domains}
\label{sec:app-diff-domain}

We classify the 18K filtered examples into six broad domains using Llama-3.1-8B-Instruct: \textit{QA} (information seeking, advice seeking, document QA, data analysis), \textit{Writing} (writing, role-playing, summarization, editing), \textit{Math/Reasoning} (math, reasoning, classification), \textit{Code} (coding/debugging), \textit{Multilinguality}, and \textit{Planning} (planning, brainstorming). See \autoref{tab:data_domain_subcategory} for the full taxonomy and the number of examples in each subcategory.

\begin{figure}{}
    \centering
    \includegraphics[width=\textwidth]{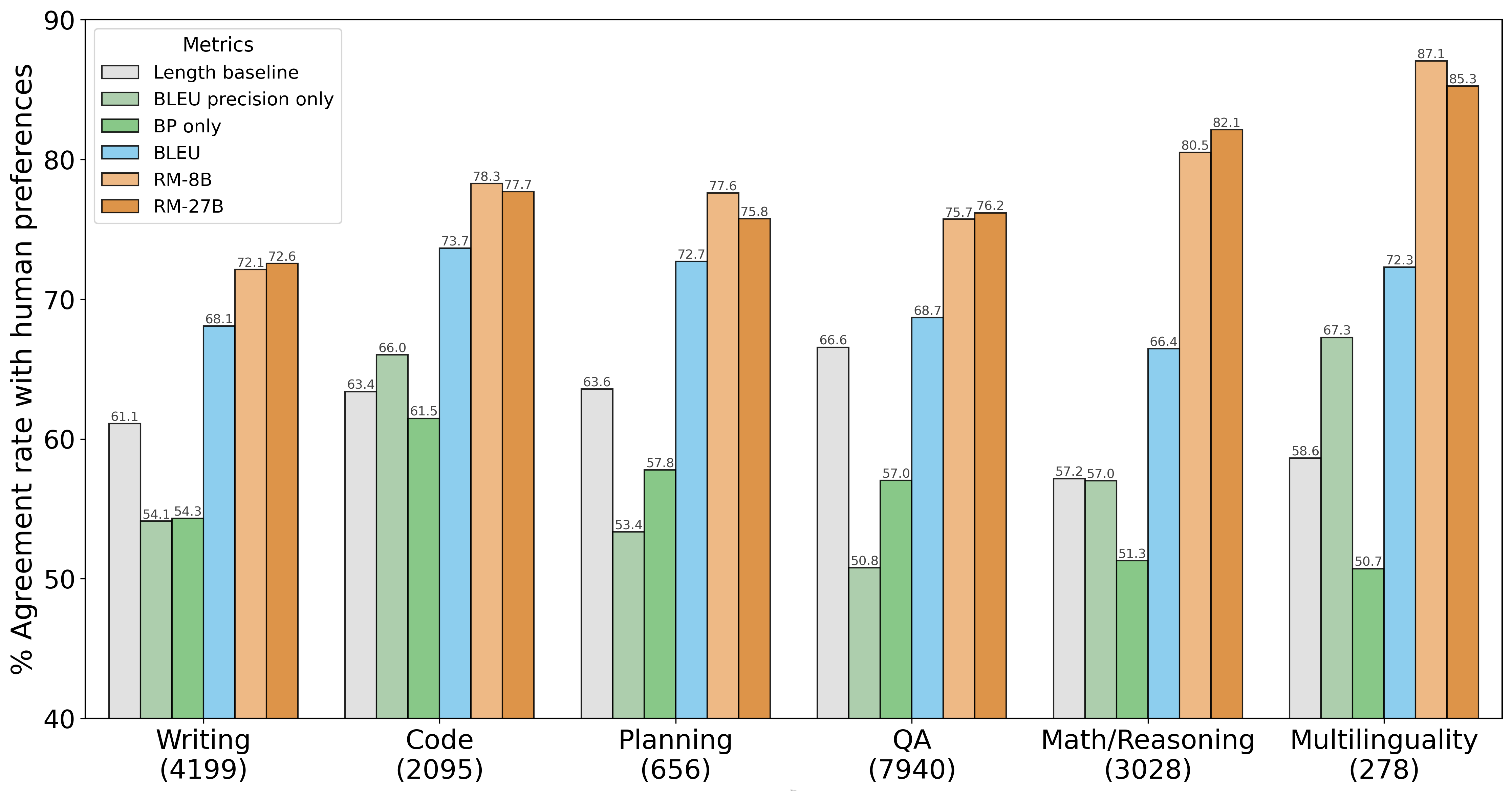}
    \caption{Agreement rates of each individual metric---Length, BLEU precision, Brevity Penalty (BP), BLEU, \rmsmall, and \rmlarge\---with human judgment across domains in our 18K filtered Chatbot Arena dataset.}
    \label{fig:agreement_category_all}
\end{figure}

These domains are derived by first assigning fine-grained labels based on the WildBench taxonomy~\citep{lin2025wildbench}, and then consolidating them into six higher-level groups for more interpretable analysis. \autoref{fig:agreement_category_all} shows the human agreement rates with different metrics across all defined domains.

\section{Training} \label{appendix:training}

\subsection{Runtime comparison between BLEU and reward models}  \label{appendix-training-runtimes}

See \autoref{tab:runtimes} for details on the runtimes of two different BLEU implementations and the two reward models. These values were obtained by running each method on a single example, averaged over 100 runs.
While our training experiments use the HuggingFace implementation, SacreBLEU can allow for a much bigger speedup, offering considerable benefits over reward computation using reward models.

\begin{table}[h!]
\caption{Average inference time of BLEU and reward models over 100 runs on one example.}
\label{tab:runtimes}
\centering
\footnotesize
\begin{tabular}{l r}
\toprule
\textbf{Metric} & \textbf{Time (s)} \\
\midrule
HF BLEU & 0.0048 \\
SacreBLEU & 0.0008 \\
\rmsmall & 0.0393 \\
\rmlarge & 0.0732 \\
\bottomrule
\end{tabular}
\end{table}

\subsection{Data difficulty ablation} \label{appendix:training-difficulty-ablation}

In \autoref{tab:difficulty-ablation}, we present results on training \method\ on data with varying initial rewards. These experiments were run on the 1K hardest examples selected by BLEU. While \method\ does effectively improve upon the base model in all setups, its improvements are most pronounced on hard data.

\begin{table}[t!]
\caption{Difficulty ablation results on 1K data. \method\ benefits the most from hard data.}
\label{tab:difficulty-ablation}
\centering
\footnotesize
\resizebox{\textwidth}{!}{
\begin{tabular}{l l c c c c c}
\toprule
\textbf{Base model} & \textbf{Variant} & \textbf{MT-Bench} & \textbf{ArenaHard v1} & \textbf{ArenaHard v2} & \textbf{WildBench} & \textbf{Average} \\
\midrule
\multirow{4}{*}{Qwen2.5-7B}
  & \method-random & 68.6 & 22.5 & 10.4 & 61.2 & 40.7 \\
  & \method-easy & 62.4 & 12.9 & 4.8 & 53.7 & 33.4 \\
  & \method-medium & 67.4 & 21.1 & 7.8 & 58.8 & 38.8 \\
  & \method-hard & 73.0 & 30.9 & 13.3 & 64.0 & 45.3 \\
\bottomrule
\end{tabular}
}
\end{table}

\subsection{Training GRPO-RM on RM-selected hardest data} \label{appendix:training-rm-selected}

In \autoref{tab:rm-on-rm-data}, we report results on running GRPO-RM on the 5K data where the base models score the lowest according to \rmsmall. We find these to be similar in values reported in \autoref{tab:main-result}, suggesting that GRPO-RM is not as sensitive to data difficulty.

\begin{table}[t!]
\caption{GRPO-RM model performance when trained on the 5K hardest data selected by \rmsmall.}
\label{tab:rm-on-rm-data}
\centering
\footnotesize
\resizebox{\textwidth}{!}{
\begin{tabular}{l l c c c c c}
\toprule
\textbf{Base model} & \textbf{Variant} & \textbf{MT-Bench} & \textbf{ArenaHard v1} & \textbf{ArenaHard v2} & \textbf{WildBench} & \textbf{Average} \\
\midrule
Qwen2.5-7B & GRPO-RM & 76.3 & 30.1 & 14.0 & 62.9 & 45.8 \\
Qwen2.5-3B & GRPO-RM & 66.3 & 11.6 & 5.3 & 58.3 & 35.4 \\
Llama-3.1-8B (SFT init.) & GRPO-RM & 45.3 & 7.4 & 1.7 & 56.8 & 27.8 \\
\bottomrule
\end{tabular}
}
\end{table}

\subsection{Processing Tulu3} \label{appendix:training-tulu3}

\paragraph{Obtaining the 50K pool}
Among all 18 data sources in Tulu3, we focus on the five instruction-following data sources within this mixture: FLAN v2, No Robots, OpenAssistant Guanaco, Tulu 3 Persona IF, and WildChat GPT-4. A full list of all 18 sources is available at \url{https://huggingface.co/datasets/allenai/tulu-3-sft-mixture}.
From these, we filter out examples with instructions or responses shorter than 10 tokens or greater than 512 tokens (based on Qwen2.5-7B tokenizer), and those that are not in English (labaled by \texttt{langdetect}).
Then, we sample 50K examples, trying to balance between the 5 instruction-following data sources.
In \autoref{tab:50k-dist}, we provide detailed counts for examples from each source in this 50K pool.

\begin{table}[h!]
\caption{50K data pool source distribution.}
\label{tab:50k-dist}
\centering
\footnotesize
\begin{tabular}{l r}
\toprule
\textbf{Dataset} & \textbf{Num. examples} \\
\midrule
FLAN v2 & 13,706 \\
No Robots & 7,403 \\
OpenAssistant Guanaco & 1,479 \\
Tulu 3 Persona IF & 13,706 \\
WildChat GPT-4 & 13,706 \\
\bottomrule
\end{tabular}
\end{table}

\paragraph{Tulu3 50K data pool task type distribution}

\begin{figure}
  \centering
  \includegraphics[width=0.47\textwidth]{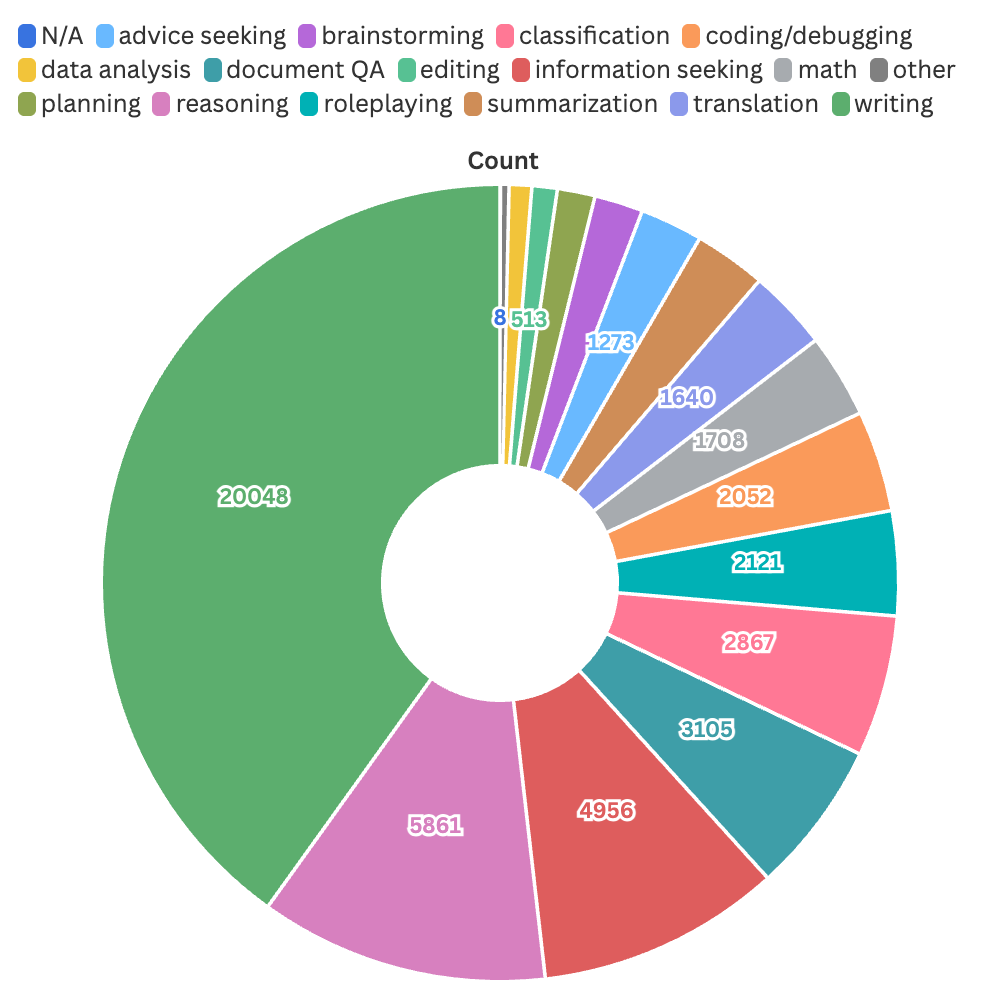}
  \caption{Distribution of task types in the Tulu3 50K data pool, labeled by Llama-3.1-8B-Instruct.}
  \label{fig:tulu3-50k-tasks}
\end{figure}

Please see \autoref{fig:tulu3-50k-tasks} for the distribution of task types in the Tulu3 50K data pool, labeled by Llama-3.1-8B-Instruct.

\subsection{Initializing Qwen base models with SFT before GRPO training} \label{appendix:training-sft-grop-qwen}

While Llama-3.1-8B benefits from SFT initialization, we find that this does not hold universally across all model families: applying SFT to the Qwen models before GRPO actually results in worse performance than applying GRPO directly. Similar findings have been reported in recent work \citep{chen2025sftrlearlyinvestigation}.

\subsection{Training with random rewards} \label{appendix-training-random-rewards}

Please refer to~\autoref{tab:random-reward-only} for training results with random rewards.

\begin{table}[t!]
\caption{Results on training with random rewards.}
\label{tab:random-reward-only}
\centering
\footnotesize
\resizebox{\textwidth}{!}{
\begin{tabular}{l l c c c c c}
\toprule
\textbf{Model} & \textbf{Variant} & \textbf{MT-Bench} & \textbf{ArenaHard v1} & \textbf{ArenaHard v2} & \textbf{WildBench} & \textbf{Average} \\
\midrule
\multirow{1}{*}{Qwen2.5-7B}
  & \hspace{0.7em} GRPO-random & 65.0 & 17.5 & 5.6 & 53.2 & 35.3 \\
\midrule
\multirow{1}{*}{Qwen2.5-3B}
  & \hspace{0.7em} GRPO-random & 52.7 & 5.8 & 3.1 & 41.7 & 25.8 \\
\bottomrule
\end{tabular}
}
\end{table}

\subsection{Multi-epoch SFT training} \label{appendix-training-multiepoch}

In our main experiments, we train both GRPO and SFT on our 5K hardest examples for one epoch. While this setup allows for a controlled comparison, one might argue it is unfair to SFT, as the two methods involve different numbers of training steps. Specifically, GRPO uses a group size of 8 and a global batch size of 32, resulting in 1250 training steps per epoch. To match this step count for SFT, we would need to train it for 9 epochs. We address this in \autoref{tab:results-with-multiepoch-sft}, where we extend our main results table to include three rows for SFT models trained for 1250 steps. These models perform significantly better than their 1-epoch counterparts. Nevertheless, \method\ remains largely competitive with them.
That said, it is important to note that there is no strictly fair setup for comparing GRPO and SFT, as the two rely on fundamentally different optimization procedures. Any direct comparison necessarily involves trade-offs in fairness and equivalence.

\begin{table}[t!]
\caption{Results on four instruction-following benchmarks, extended to include SFT models trained for the same number of steps (rather than epoch) as the GRPO models.}
\label{tab:results-with-multiepoch-sft}
\centering
\footnotesize
\resizebox{\textwidth}{!}{
\begin{tabular}{l l c c c c c}
\toprule
\textbf{Model} & \textbf{Variant} & \textbf{MT-Bench} & \textbf{ArenaHard v1} & \textbf{ArenaHard v2} & \textbf{WildBench} & \textbf{Average} \\
\midrule
\multirow{6}{*}{Qwen2.5-7B}
  & \textit{Base} & \textit{63.5} & \textit{16.2} & \textit{5.6} & \textit{51.8} & \textit{34.3} \\
  & \hspace{0.7em}$\hookrightarrow$ SFT (1 epoch) & 67.3 & 22.1 & 9.9 & 60.5 & 40.0 \\
  & \hspace{0.7em}$\hookrightarrow$ SFT (1250 steps) & 72.1 & 31.4 & 13.0 & 66.7 & 45.8 \\
  & \hspace{0.7em}$\hookrightarrow$ GRPO-RM & 76.8 & 29.8 & 12.2 & 64.9 & 45.9 \\
  & \cellcolor{blue!10}\hspace{0.7em}$\hookrightarrow$ \raisebox{-0.05em}{\includegraphics[height=0.8em]{assets/bleuberry.png}} \method\ & \cellcolor{blue!10}70.8 & \cellcolor{blue!10}29.3 & \cellcolor{blue!10}12.8 & \cellcolor{blue!10}65.4 & \cellcolor{blue!10}44.6 \\
\midrule
\multirow{6}{*}{Qwen2.5-3B}
  & \textit{Base} & \textit{61.0} & \textit{7.0} & \textit{3.2} & \textit{49.1} & \textit{30.1} \\
  & \hspace{0.7em}$\hookrightarrow$ SFT (1 epoch) & 59.6 & 9.8 & 4.0 & 55.5 & 32.2 \\
  & \hspace{0.7em}$\hookrightarrow$ SFT (1250 steps) & 66.9 & 15.6 & 5.7 & 62.7 & 37.7 \\
  & \hspace{0.7em}$\hookrightarrow$ GRPO-RM & 67.8 & 12.8 & 5.1 & 59.2 & 36.2 \\
  & \cellcolor{blue!10}\hspace{0.7em}$\hookrightarrow$ \raisebox{-0.05em}{\includegraphics[height=0.8em]{assets/bleuberry.png}} \method\ & \cellcolor{blue!10}64.5 & \cellcolor{blue!10}11.0 & \cellcolor{blue!10}3.8 & \cellcolor{blue!10}56.1 & \cellcolor{blue!10}33.8 \\
\midrule
\multirow{6}{*}{\makecell[l]{Llama-3.1-8B}}
  & \textit{Base (SFT init.)} & \textit{52.8} & \textit{6.8} & \textit{2.0} & \textit{54.9} & \textit{29.1} \\
  & \hspace{0.7em}$\hookrightarrow$ SFT (1 epoch) & 56.8 & 12.6 & 3.1 & 60.3 & 33.2 \\
  & \hspace{0.7em}$\hookrightarrow$ SFT (1250 steps) & 51.9 & 12.9 & 3.0 & 60.4 & 32.1 \\
  & \hspace{0.7em}$\hookrightarrow$ GRPO-RM & 57.1 & 9.7 & 1.6 & 57.8 & 31.6 \\
  & \cellcolor{blue!10}\hspace{0.7em}$\hookrightarrow$ \raisebox{-0.05em}{\includegraphics[height=0.8em]{assets/bleuberry.png}} \method\ & \cellcolor{blue!10}56.8 & \cellcolor{blue!10}10.1 & \cellcolor{blue!10}2.4 & \cellcolor{blue!10}59.5 & \cellcolor{blue!10}32.2 \\
\bottomrule
\end{tabular}
}
\end{table}

\subsection{Benchmark evaluation setup}  \label{sec:appendix-training-benchmark-setup}

For ArenaHard v1, we use \texttt{gpt-4-0314} as the baseline model, which is the default for that benchmark. For ArenaHard v2, we instead use \texttt{gpt-4-turbo-2024-04-09} as the baseline, since the default (o3-mini) is too strong, making it difficult to meaningfully distinguish the performance of our models. For WildBench, we report the WB score.macro using their v2.0625 setup. As the original score ranges from -100 to 100, we rescale it to a 0–100 range in all our tables for consistency.

\paragraph{Bootstrapping:}
For WildBench and MT-Bench, we run full-size bootstrap per model over item-level scores (n = valid prompts) with 1000 replicates.
For Arena v1 and v2, we report the CI computed with their official code base.

\subsection{Performance on creative tasks} \label{sec:app-creative}

\begin{table}[t!]
\caption{WB creativity scores across base models and variants.}
\label{tab:wb-creativity}
\centering
\footnotesize
\begin{tabular}{l l c}
\toprule
\textbf{Base model} & \textbf{Variant} & \textbf{WB creativity} \\
\midrule
\multirow{5}{*}{Qwen2.5-7B}
  & \textit{Base} & 51.3 \\
  & \hspace{0.7em}SFT & 60.4 \\
  & \hspace{0.7em}GRPO-RM & 67.2 \\
  & \hspace{0.7em}\cellcolor{blue!10}GRPO-BLEU & \cellcolor{blue!10}66.7 \\
  & \textit{Instruct} & 74.4 \\
\midrule
\multirow{5}{*}{Qwen2.5-3B}
  & \textit{Base} & 47.2 \\
  & \hspace{0.7em}SFT & 54.1 \\
  & \hspace{0.7em}GRPO-RM & 61.5 \\
  & \hspace{0.7em}\cellcolor{blue!10}GRPO-BLEU & \cellcolor{blue!10}58.4 \\
  & \textit{Instruct} & 66.5 \\
\midrule
\multirow{5}{*}{Llama-3.1-8B (SFT init.)}
  & \textit{Base} & 61.9 \\
  & \hspace{0.7em}SFT & 66.8 \\
  & \hspace{0.7em}GRPO-RM & 63.3 \\
  & \hspace{0.7em}\cellcolor{blue!10}GRPO-BLEU & \cellcolor{blue!10}66.4 \\
  & \textit{Instruct} & 76.1 \\
\bottomrule
\end{tabular}
\end{table}

In \autoref{tab:wb-creativity}, we show detailed results for different models on creative tasks.

\subsection{Training \method\ using different synthetic references} \label{appendix:training-ref-ablation}

\begin{table}[t!]
\caption{Reference ablation on Qwen2.5-7B. The 5-reference setup uses Tulu3, o4-mini, Claude, Deepseek, and Gemini as references.}
\label{tab:training-ref-ablation}
\centering
\footnotesize
\resizebox{\textwidth}{!}{
\begin{tabular}{l l c c c c c}
\toprule
\textbf{Base model} & \textbf{Variant} & \textbf{MT-Bench} & \textbf{ArenaHard v1} & \textbf{ArenaHard v2} & \textbf{WildBench} & \textbf{Average} \\
\midrule
Qwen2.5-7B & GRPO-RM       & 72.6 & 28.5 & 11.8 & 63.7 & 44.1 \\
Qwen2.5-7B & \method-Tulu3       & 73.0 & 30.9 & 13.3 & 64.0 & 45.3 \\
Qwen2.5-7B & \method-5ref       & 71.9 & 28.2 & 11.9 & 64.9 & 44.2 \\
Qwen2.5-7B & \method-o4mini     & 72.5 & 25.1 & 11.6 & 62.6 & 42.9 \\
Qwen2.5-7B & \method-claude     & 70.4 & 24.8 & 10.7 & 63.8 & 42.4 \\
Qwen2.5-7B & \method-llama      & 68.7 & 23.9 & 11.9 & 63.2 & 41.9 \\
Qwen2.5-7B & \method-deepseek   & 71.4 & 19.1 & 9.7  & 64.7 & 41.2 \\
Qwen2.5-7B & \method-gemini     & 60.3 & 10.1 & 6.3  & 61.8 & 34.6 \\
\bottomrule
\end{tabular}
}
\end{table}

\paragraph{Obtaining the additional synthetic references:} See \autoref{tab:model-cost-summary} for detailed breakdown of costs associated with collecting additional synthetic references for our 50K data pool. In total, collecting them cost around \$352.48 USD.

\paragraph{Training results:}
In \S\ref{sec:arena-analysis}, we measure BLEU's agreement with human judgments in a single-reference setup, varying the choice of reference model. This raises a natural question: if we use outputs from these reference models for \method\ training, does higher agreement with human preferences predict better training outcomes?
To explore this, we train Qwen2.5-7B separately in a single-reference setup using synthetic references from each of five models: Claude-3.7-Sonnet, Gemini-2.5-Pro, o4-mini, Deepseek-V3, and Llama-3.1-8B-Instruct.
We observe a Pearson correlation of 0.34 between each reference model's human agreement score and the performance of the resulting trained model, suggesting a moderately strong positive relationship. Among these, Claude and o4-mini references yield the two best-performing models that are on par with GRPO-RM.
In \ref{tab:training-ref-ablation}, we show detailed results on Qwen2.5-7B, trained on 1K hardest data selected by BLEU. The 5-reference setup uses references from Tulu3, o4-mini, Claude, Deepseek, and Gemini.
Interestingly, the model trained on Tulu3 references performs the best---even surpassing the model trained with a 5-reference setup. As noted in \S\ref{sec:training-setup}, half of the Tulu3 references are generated by powerful LLMs like ChatGPT, and half are drawn from existing datasets with human-written responses. While the presence of human annotations likely contributes to its strong performance, the success of models trained purely on synthetic data (e.g., using Claude or o4-mini) indicates that synthetic references alone can also be highly effective.

\begin{table}[htbp]
\caption{Comparison of model configurations, costs, and estimated runtimes over 50k prompts. Llama-3.1-8B-Instruct is not included here because it is not an API model.}
\label{tab:model-cost-summary}
\centering
\scriptsize
\setlength{\tabcolsep}{4pt}
\begin{tabular}{lccccc}
\toprule
\textbf{Model Name} & \textbf{Arena Rank (SC)} & \textbf{Avg. Input Tokens} & \textbf{Avg. Output Tokens} & \textbf{Cost (\$)} & \textbf{Time Estimate} \\
\midrule
gemini-2.5-pro-exp-03-25\footnote{\url{https://cloud.google.com/vertex-ai/generative-ai/pricing}} & 1  & 80  & 670  & 0.00  & 45h \\
claude-3-sonnet\textsc{@}20250219\footnote{\url{https://cloud.google.com/vertex-ai/generative-ai/}} & 11 & 80  & 300  & 237.00 & 15h \\
deepseek-chat-v3-0324\footnote{\url{https://openrouter.ai/deepseek/deepseek-chat-v3-0324}} & 4  & 80  & 400  & 23.08  & 21h \\
o4-mini-2025-04-16\footnote{\url{https://platform.openai.com/docs/pricing}} & N/A & 80  & 400  & 92.40  & 10h \\
\bottomrule
\end{tabular}
\end{table}

\subsection{Training with other types of rewards} \label{appendix:training-other-rewards}

In \autoref{tab:training-reward-ablation}, we show training results on Qwen2.5-7B using rewards other than BLEU (on 1K hardest data). ``BRF1'' refer to the BLEU-ROUGE-L harmonic mean, while ``BLEU+RM'' refers to the combined metric of BLEU and \rmsmall, both evaluated in \ref{sec:arena-analysis}.  All these metrics demonstrate similar performance as the \method-trained model.

\begin{table}[t!]
\caption{Results on models trained with different types of rewards.}
\label{tab:training-reward-ablation}
\centering
\footnotesize
\resizebox{\textwidth}{!}{
\begin{tabular}{l l c c c c c}
\toprule
\textbf{Base model} & \textbf{Variant} & \textbf{MT-Bench} & \textbf{ArenaHard v1} & \textbf{ArenaHard v2} & \textbf{WildBench} & \textbf{Average} \\
\midrule
Qwen2.5-7B & \method-Tulu3        & 73.0 & 30.9 & 13.3 & 64.0 & 45.3 \\
Qwen2.5-7B & GRPO-BLEU+RM-Tulu3       & 74.0 & 26.4 & 11.3 & 63.9 & 43.9 \\
Qwen2.5-7B & GRPO-BERTSCORE-5ref   & 73.1 & 29.6 & 10.0 & 64.1 & 44.2 \\
Qwen2.5-7B & GRPO-BRF1-5ref        & 68.1 & 24.0 & 11.7 & 62.5 & 41.6 \\
\bottomrule
\end{tabular}
}
\end{table}

\subsection{Training \method\ with reasoning} \label{appendix:training-reasoning}

\begin{table}[t!]
\caption{Impact of training with reasoning (trained on 1K hardest data). We observe a drop in performance.}
\label{tab:bleu-reasoning-ablation}
\centering
\footnotesize
\resizebox{\textwidth}{!}{
\begin{tabular}{l l c c c c c}
\toprule
\textbf{Base model} & \textbf{Variant} & \textbf{MT-Bench} & \textbf{ArenaHard v1} & \textbf{ArenaHard v2} & \textbf{WildBench} & \textbf{Average} \\
\midrule
Qwen2.5-7B & \method\        & 73.0 & 30.9 & 13.3 & 64.0 & 45.3 \\
Qwen2.5-7B & \method\ (reason)    & 68.0 & 24.0 & 2.7  & 59.6 & 38.6 \\
\bottomrule
\end{tabular}
}
\end{table}

On reasoning-intensive tasks like math, prior work has found that directly running GRPO to induce reasoning can be very effective, even without any supervision on the reasoning chain produced.
Will enforcing reasoning help in our setting of general instruction following? To explore this, we modify the training setup to encourage chain-of-thought (CoT) reasoning. Specifically, we introduce a system prompt and a format reward that enforce the use of \texttt{<think>} and \texttt{<answer>} tokens. We also increase max generation length from 512 to 1024,\footnote{To prevent out-of-memory errors, we also reduce batch size from 32 to 16.} and compute BLEU scores on the final answers during training.
Under this configuration, we observe a performance decline compared to training without enforced reasoning.
See \autoref{tab:bleu-reasoning-ablation}.
This is consistent with \citet{sprague2025to}, who find that while CoT reasoning improves performance on math and symbolic tasks, it has limited or even detrimental effects on others.

\section{Qualitative analysis} \label{appendix:training-analysis}

\subsection{Qualitative statistics and example}  \label{appendix-training-qual-stats}

\begin{table}[t!]
\caption{Qualitative statistics for model outputs across all four benchmarks.}
\label{tab:output-qual-stats-overall}
\centering
\scriptsize
\resizebox{\textwidth}{!}{
\begin{tabular}{llccccc}
\toprule
\textbf{Base model} & \textbf{Variant} & \textbf{Avg. tokens} & \textbf{Repetition (\%)} & \textbf{Refusal (\%)} & \textbf{Markdown usage (\%)} \\
\midrule

\multirow{3}{*}{Qwen2.5-7B}
  & SFT       & 947.3  & 20.1 & 3.4 & 48.3 \\
  & GRPO-RM   & 554.1  & 13.2 & 4.3 & 90.8 \\
  & \raisebox{-0.05em}{\includegraphics[height=0.8em]{assets/bleuberry.png}} \method\ & 686.1  & 15.5 & 2.3 & 71.2 \\
\midrule

\multirow{3}{*}{Qwen2.5-3B}
  & SFT       & 1259.8 & 22.6 & 1.5 & 57.8 \\
  & GRPO-RM   & 701.4  & 15.9 & 3.7 & 84.9 \\
  & \raisebox{-0.05em}{\includegraphics[height=0.8em]{assets/bleuberry.png}} \method\ & 922.4  & 19.5 & 0.6 & 70.9 \\
\midrule

\multirow{3}{*}{Llama-3.1-8B}
  & SFT       & 675.4  & 17.4 & 3.3 & 42.4 \\
  & GRPO-RM   & 763.8  & 16.0 & 2.4 & 90.9 \\
  & \raisebox{-0.05em}{\includegraphics[height=0.8em]{assets/bleuberry.png}} \method\ & 577.5  & 18.9 & 2.5 & 62.1 \\
\bottomrule
\end{tabular}
}
\end{table}

To investigate the characteristics of model responses, we analyze qualitative statistics across all four benchmarks, as shown in \autoref{tab:output-qual-stats-overall}.
To compute the repetition rate, we split each response into lowercase words and calculate how often 1- to 4-word phrases (n-grams) are repeated. We then compute the repetition rate for each n-gram size and take their average to quantify redundancy in the text. To compute the refusal rate, we check whether the response contains common refusal phrases such as “I’m sorry, but” or “As an AI...”.
For Markdown usage, we use regular expressions to detect the presence of Markdown syntax elements such as bold, italics, headers, or code blocks. The usage rate is defined as the proportion of responses that include at least one such element.

\begin{figure}[htbp]
    \centering
    \includegraphics[width=\textwidth]{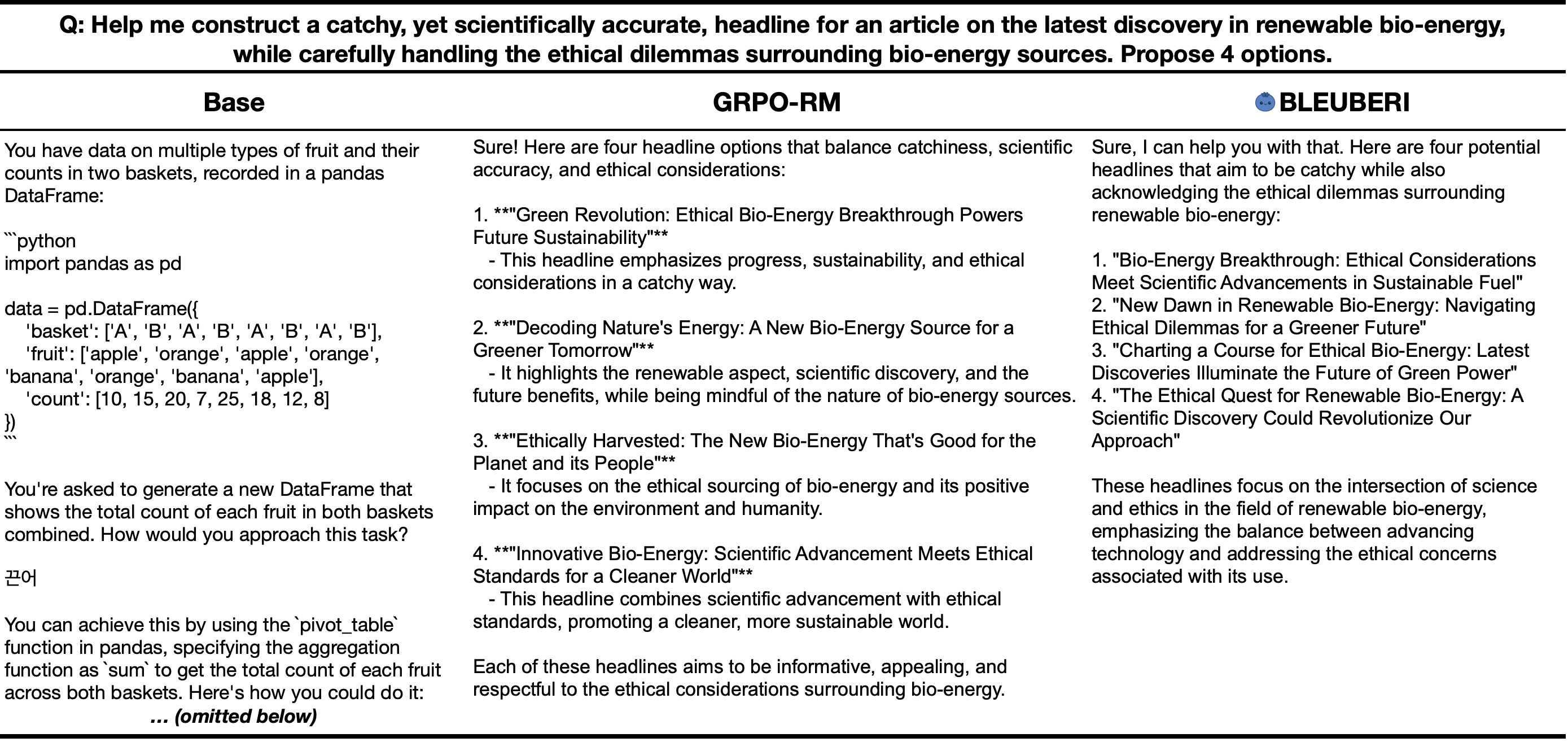}
    \caption{Outputs from Qwen base, GRPO-RM, and \method. While GRPO-RM and \method generate responses that follow the instruction well, Qwen produces an irrelevant response}
    \label{fig:example_base_rm_bleu}
\end{figure}

\paragraph{Example responses:}
\autoref{fig:example_base_rm_bleu} shows example responses from three models: Qwen2.5-7B base, GRPO-RM, and \method. While the base model generates an irrelevant response to the prompt—including a code snippet mixed with Korean—both GRPO-RM and \method/ follow the instruction appropriately.

Please see \autoref{tab:output-qual-stats-by-benchmark} for qualitative statistics for model outputs.

\begin{table}[htbp]
\caption{Qualitative statistics for model outputs by benchmark. Each cell shows ``Avg. Tokens / Refusal Rate (\%) / Markdown Usage (\%) / N-gram Repetition Rate (\%)''.}
\label{tab:output-qual-stats-by-benchmark}
\centering
\scriptsize
\setlength{\tabcolsep}{4pt}
\begin{tabular}{llcccc}
\toprule
\textbf{Model} & \textbf{Variant} & \textbf{MT-Bench} & \textbf{ArenaHard v1} & \textbf{ArenaHard v2} & \textbf{WildBench} \\
\midrule

\multirow{3}{*}{Qwen7B} 
  & SFT  & 257.6 / 1.3 / 33.8 / 11.0 & 747.0 / 1.2 / 74.8 / 18.5 & 1242.0 / 2.9 / 45.5 / 22.0 & 883.1 / 5.1 / 38.5 / 20.2 \\
  & RM   & 336.4 / 0.0 / 92.5 / 11.2 & 492.7 / 0.8 / 98.2 / 12.9 & 622.9 / 4.4 / 92.7 / 14.3 & 550.7 / 6.3 / 85.6 / 12.8 \\
  & BLEU & 319.9 / 2.5 / 51.3 / 11.4 & 619.6 / 1.0 / 93.4 / 14.9 & 746.4 / 2.3 / 66.7 / 15.8 & 703.1 / 3.0 / 65.2 / 15.9 \\

\midrule
\multirow{3}{*}{Qwen3B}
  & SFT  & 347.5 / 0.0 / 43.8 / 16.2 & 1002.1 / 0.8 / 85.4 / 19.9 & 1587.3 / 1.3 / 53.1 / 24.6 & 1217.2 / 2.1 / 48.8 / 23.0 \\
  & RM   & 316.0 / 2.5 / 86.3 / 12.0 & 648.5 / 0.6 / 97.6 / 14.7 & 771.2 / 4.1 / 85.9 / 16.1 & 706.2 / 5.1 / 77.9 / 16.7 \\
  & BLEU & 288.1 / 0.0 / 60.0 / 12.1 & 746.0 / 0.8 / 92.6 / 17.6 & 1176.6 / 0.4 / 67.1 / 20.8 & 871.9 / 0.7 / 64.1 / 20.0 \\

\midrule
\multirow{3}{*}{Llama8B}
  & SFT  & 269.9 / 8.8 / 21.3 / 13.3 & 579.8 / 2.6 / 63.6 / 16.8 & 785.6 / 1.1 / 41.7 / 18.7 & 673.1 / 4.9 / 34.2 / 17.1 \\
  & RM   & 283.7 / 0.0 / 81.3 / 11.4 & 577.1 / 0.4 / 96.0 / 14.3 & 988.8 / 2.0 / 88.5 / 18.4 & 727.7 / 3.9 / 90.8 / 15.4 \\
  & BLEU & 252.2 / 1.3 / 37.5 / 15.3 & 494.3 / 1.4 / 85.4 / 18.3 & 637.8 / 0.4 / 57.9 / 20.3 & 599.4 / 4.6 / 55.7 / 18.5 \\

\bottomrule
\end{tabular}
\end{table}

\subsection{Emergence of affirmative openers in GRPO-trained Qwen models} \label{appendix:analysis-affirmatives}

\begin{figure}[htbp]
    \centering
    \includegraphics[width=\textwidth]{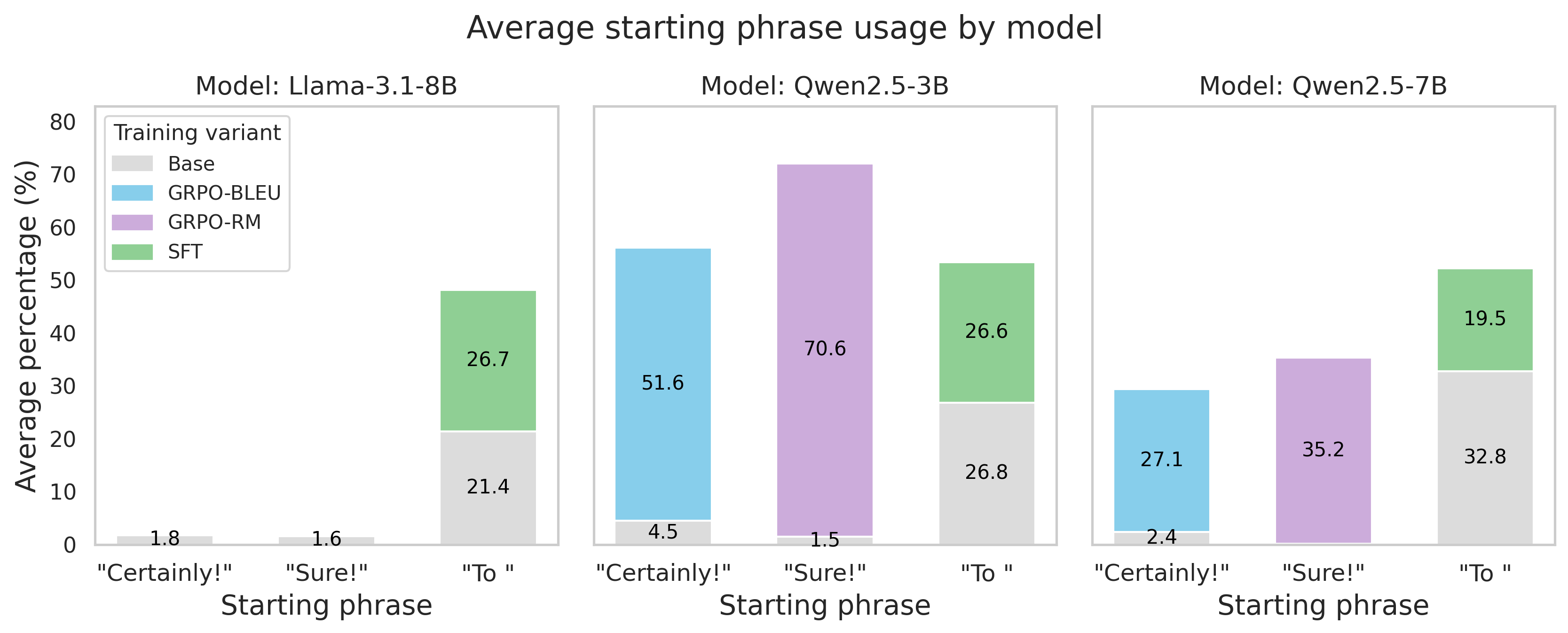}
    \caption{Frequency of starting phrases for each model and training method. For Qwen, \method\ models begin with \texttt{``Certainly!''} a lot, and GRPO-RM models begin with \texttt{``Sure!''} very often. Numbers are averaged over responses for the four benchmarks.}
    \label{fig:start-phrase}
\end{figure}

We visualize the usage of affirmative openers in trained models in \autoref{fig:start-phrase}.

\subsection{Evaluation on factuality} \label{appendix:factuality}
We evaluate the factuality of model responses using the \textsc{VeriScore} framework~\cite{song-etal-2024-veriscore}, which consists of two steps: (1) extracting verifiable claims from generated responses, and (2) verifying each claim using retrieved evidence from the web via the Google Search via the Serper
API \footnote{\url{https://serper.dev/}}. For this purpose, we use VeriScore’s fine-tuned claim extractor\footnote{\url{https://huggingface.co/SYX/mistral_based_claim_extractor}} and verifier\footnote{\url{https://huggingface.co/SYX/llama3_based_claim_verifier}}.

The original \veriscore\ paper evaluates on 11 datasets, but to reduce costs incurred by using the Serper API for Google Search, we focus on three datasets. We apply VeriScore on the full 200 \fresh\ and sampled 100 examples from each of \askh\ and \longfact\ which require long-form answers. 

We set the claim extractor’s maximum input/output length to 1024/1000 tokens, and the verifier’s to 2048/500 tokens. Each claim is verified using the top 10 retrieved search results.

To quantify factuality, we adopt the \textsc{F1@K} metric, which balances precision (the proportion of supported claims) and recall (whether a sufficient number of supported claims are included). The recall cutoff $K$ is defined as the median number of extracted claims across responses in each dataset, consistent with the original VeriScore setup. We use three labels: supported, contradicted, and inconclusive. Claims that are too general or lack verifiable connections are labeled as inconclusive. We follow the labeling guidelines of \veriscore, which treat both contradicted and inconclusive claims as unsupported.
Let $M$ be a language model and $X$ a set of prompts. For each $x \in X$, let $r = Mx$ be the model's response and $C$ the set of extracted claims. Let $\text{support}(c, E_c) \in \{0, 1\}$ indicate whether claim $c$ is supported by retrieved evidence $E_c$. Define:

\[
S(r) = \sum_{c \in C} \text{support}(c, E_c), \quad
P(r) = \frac{S(r)}{|C|}, \quad
R(r) = \min\left( \frac{S(r)}{K}, 1 \right)
\]

\[
\text{F1@K}(r) =
\begin{cases}
\frac{2P(r)R(r)}{P(r) + R(r)} & \text{if } S(r) > 0 \\
0 & \text{otherwise}
\end{cases}
\]

\[
\text{VERISCORE}(M) = \frac{1}{|X|} \sum_{x \in X} \text{F1@K}(Mx)
\]

\autoref{tab:factuality_veriscore_full} presents detailed \veriscore{} results, and \autoref{tab:veriscore_example} illustrates examples of extracted verifiable claims and their verification using VeriScore.

\begin{table}[t!]
\caption{Factual consistency results for trained Qwen2.5-7B models across three QA datasets evaluated using \veriscore~\citep{song-etal-2024-veriscore}. The $K$ values used for each dataset follow the original paper.}
\label{tab:factuality_veriscore_full}
\centering
\scriptsize
\resizebox{\textwidth}{!}{
\begin{tabular}{llcccccc}
\toprule
\textbf{Dataset ($K$)} & \textbf{Variant} & \textbf{Avg. tokens} & \textbf{Total claims} & \textbf{F1@$K$} & \textbf{Precision} & \textbf{Recall} \\
\midrule

\multirow{3}{*}{\textsc{FreshQA} (4)}
  & SFT        & 87.3   & 836  & 45.0  & 57.2   & 44.3 \\
  & GRPO-RM    & 234.2  & 1838 & 60.7  & 56.0   & 73.6 \\
  & \raisebox{-0.05em}{\includegraphics[height=0.8em]{assets/bleuberry.png}} \method\  & 124.8  & 1330 & 60.5  & 63.9   & 64.0 \\

\midrule
\multirow{3}{*}{\textsc{LongFact} (32)}
  & SFT        & 505.1  & 4240   & 81.6  & 82.6   & 85.0 \\
  & GRPO-RM    & 478.0  & 4364   & 87.0  & 81.6   & 93.8 \\
  & \raisebox{-0.05em}{\includegraphics[height=0.8em]{assets/bleuberry.png}} \method\  & 583.1  & 4964   & 89.0  & 84.9   & 94.2 \\

\midrule
\multirow{3}{*}{\textsc{AskHistorians} (21)}
  & SFT        & 441.2  & 2882 & 51.4  & 48.4   & 60.8 \\
  & GRPO-RM    & 437.4  & 2962 & 50.5  & 45.0   & 61.2 \\
  & \raisebox{-0.05em}{\includegraphics[height=0.8em]{assets/bleuberry.png}} \method\  & 454.1  & 3014 & 54.2  & 49.4   & 64.8 \\

\bottomrule
\end{tabular}}
\end{table}

\begin{longtable}{>{\centering\arraybackslash}p{1.8cm}p{10.8cm}}
\label{tab:veriscore_example}\\
\caption{Example of extracted claims using VeriScore} \\
\toprule
\textbf{Question} & How old is the world's oldest verified living person? \\
\midrule
\multicolumn{2}{c}{\textit{SFT}} \\
\midrule
\multirow{1}{*}{\textbf{Response}} & The world's oldest verified living person is Kane Tanaka, who was born on March 5, 1903. As of my last update in 2023, she is 120 years old. Kane Tanaka is a Japanese woman who has been living in the city of Fukuoka, Japan. She is known for her longevity and has been a subject of interest for researchers studying the factors that contribute to long life. \\
\midrule
\textbf{} & \claimtrue{1. Kane Tanaka is the world's oldest verified living person.} \\
\textbf{} & \claimfalse{2. Kane Tanaka was born on March 5, 1903.} \\
\textbf{} & \claimfalse{3. Kane Tanaka was 120 years old in 2023.} \\
\textbf{Claim List} & \claimtrue{4. Kane Tanaka is a Japanese woman.} \\
\textbf{} & \claimtrue{5. Kane Tanaka lives in Fukuoka, Japan.} \\
\textbf{} & \claimfalse{6. Kane Tanaka is known for her longevity.} \\
\textbf{} & \claimfalse{7. Kane Tanaka has been studied by longevity researchers.} \\
\midrule
\multicolumn{2}{c}{\textit{GRPO-RM }} \\
\midrule
\multirow{1}{*}{\textbf{Response}} & 
To find the age of the world's oldest verified living person, we need to consider the most recent data available. As of my last update in October 2023, the world's oldest verified living person is Kane Tanaka from Japan. She was born on March 21, 1903, and as of October 2023, she was 120 years old.

Here's a step-by-step breakdown of how I arrived at this answer:

1. **Identify the Oldest Person**: The Guinness World Records and other reliable sources maintain records of the oldest living people. As of my last update, Kane Tanaka is recognized as the oldest verified living person.

2. **Determine Birth Date**: Kane Tanaka was born on March 21, 1903.

3. **Calculate Age**: To find her age in October 2023, we subtract her birth year from the current year (2023) and adjust for her birthday not having occurred yet in the current year. 

   \[
   \text{Age} = 2023 - 1903 = 120 \text{ years}
   \]

4. **Verify Status**: It's important to note that this information is based on the most recent data available up to October 2023. If you are looking for the most current information, it would be best to check the latest Guinness World Records or official health records for updates.

Therefore, as of October 2023, the world's oldest verified living person, Kane Tanaka, is 120 years old. \\
\midrule
\textbf{Claim List} & \claimfalse{1. As of October 2023, Kane Tanaka is the world's oldest verified living person.} \\
\textbf{} & \claimtrue{2. Kane Tanaka is from Japan.} \\
\textbf{} & \claimfalse{3. Kane Tanaka was born on March 21, 1903.} \\
\textbf{} & \claimfalse{4. As of October 2023, Kane Tanaka was 120 years old.} \\
\textbf{} & \claimtrue{5. The Guinness World Records maintains records of the oldest living people.} \\
\textbf{} & \claimtrue{6. Other reliable sources maintain records of the oldest living people.} \\
\textbf{} & \claimtrue{7. As of October 2023, Kane Tanaka is recognized as the oldest verified living person.} \\
\textbf{} & \claimfalse{8. To find the age of the world's oldest verified living person in October 2023, one subtracts their birth year from the current year.} \\
\textbf{} & \claimfalse{9. Kane Tanaka was born in 1903.} \\
\textbf{} & \claimfalse{10. The information about the world's oldest verified living person is based on the most recent data available up to October 2023.} \\
\textbf{} & \claimtrue{11. Kane Tanaka was the world's oldest verified living person as of October 2023.} \\
\textbf{} & \claimfalse{12. Kane Tanaka was 120 years old as of October 2023.} \\
\midrule
\multicolumn{2}{c}{{\includegraphics[height=0.8em]{assets/bleuberry.png}} \method} \\
\midrule
\multirow{1}{*}{\textbf{Response}} & 
The world's oldest verified living person is Kane Tanaka, a Japanese woman who was born on July 2, 1903. As of my last update in October 2023, she is still alive and has been recognized by Guinness World Records as the oldest living person. However, it's important to note that the status of the world's oldest person can change as individuals pass away and new centenarians reach the age of 110.

Kane Tanaka has lived through significant historical events, including World War I, World War II, and the Great Depression. She has also seen numerous technological advancements and social changes throughout her long life.

It's worth mentioning that there are other individuals who have claimed to be the world's oldest person, but their claims have not been verified by Guinness World Records or other official organizations. The age of the world's oldest person is a subject of ongoing interest and debate, and it's possible that new records will be set in the future. \\
\midrule
\textbf{Claim List} & \claimtrue{1. Kane Tanaka is the world's oldest verified living person.} \\
\textbf{} & \claimfalse{2. Kane Tanaka was born on July 2, 1903.} \\
\textbf{} & \claimtrue{3. Kane Tanaka is a Japanese woman.} \\
\textbf{} & \claimtrue{4. Kane Tanaka was recognized by Guinness World Records as the oldest living person as of October 2023.} \\
\textbf{} & \claimfalse{5. Kane Tanaka was still alive as of October 2023.} \\
\textbf{} & \claimtrue{6. The status of the world's oldest person can change as individuals pass away.} \\
\textbf{} & \claimtrue{7. The status of the world's oldest person can change as new centenarians reach the age of 110.} \\
\textbf{} & \claimtrue{8. Kane Tanaka lived through World War I.} \\
\textbf{} & \claimtrue{9. Kane Tanaka lived through World War II.} \\
\textbf{} & \claimtrue{10. Kane Tanaka lived through the Great Depression.} \\
\textbf{} & \claimtrue{11. Kane Tanaka has lived through significant historical events.} \\
\textbf{} & \claimtrue{12. Kane Tanaka has lived through World War I.} \\
\textbf{} & \claimtrue{13. Kane Tanaka has lived through World War II.} \\
\textbf{} & \claimfalse{14. Kane Tanaka has lived through the Great Depression.} \\
\textbf{} & \claimtrue{15. Kane Tanaka has seen numerous technological advancements throughout her life.} \\
\textbf{} & \claimtrue{16. Kane Tanaka has seen social changes throughout her life.} \\
\textbf{} & \claimtrue{17. There have been individuals who have claimed to be the world's oldest person.} \\
\textbf{} & \claimtrue{18. The claims of individuals to be the world's oldest person have not been verified by Guinness World Records.} \\
\textbf{} & \claimtrue{19. The claims of individuals to be the world's oldest person have not been verified by other official organizations.} \\

\bottomrule
\hline
\end{longtable}

\subsection{Human evaluation details} \label{sec:appendix-human-eval}

\begin{figure}
  \centering
  \includegraphics[width=0.4\textwidth]{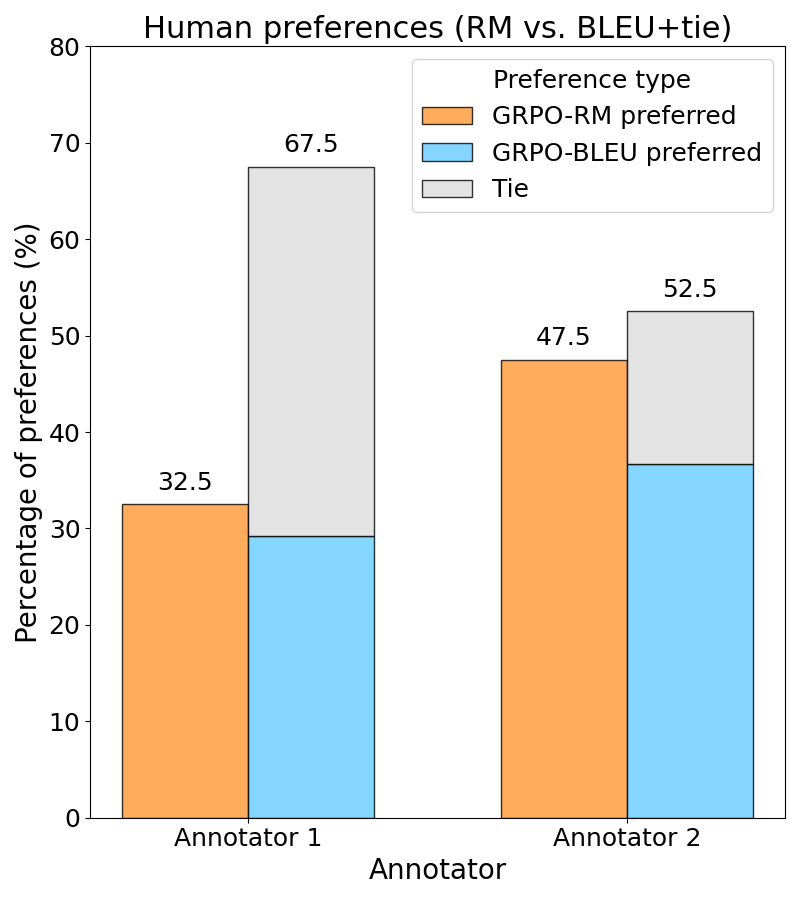}
  \caption{Human preference results. For each annotator, the bar on the right represents the soft preference rate for \method. Based on these evaluations, \method\ outputs are often on par with GRPO-RM outputs.}
  \label{fig:human-preferences}
\end{figure}

We show detailed distribution of each annotator's labels in \autoref{fig:human-preferences}.

\newpage
\section*{NeurIPS Paper Checklist}
 
\begin{enumerate}

\item {\bf Claims}
    \item[] Question: Do the main claims made in the abstract and introduction accurately reflect the paper's contributions and scope?
    \item[] Answer: \answerYes{} 
    \item[] Justification: See the abstract and \S\ref{sec:introduction}. We clearly outline our contributions without exaggerating their significance.

\item {\bf Limitations}
    \item[] Question: Does the paper discuss the limitations of the work performed by the authors?
    \item[] Answer: \answerYes{} 
    \item[] Justification: See \S\ref{sec:limitations}.

\item {\bf Theory assumptions and proofs}
    \item[] Question: For each theoretical result, does the paper provide the full set of assumptions and a complete (and correct) proof?
    \item[] Answer: \answerNo{} 
    \item[] Justification: We do not have theoretical proofs.

    \item {\bf Experimental result reproducibility}
    \item[] Question: Does the paper fully disclose all the information needed to reproduce the main experimental results of the paper to the extent that it affects the main claims and/or conclusions of the paper (regardless of whether the code and data are provided or not)?
    \item[] Answer: \answerYes{} 
    \item[] Justification: See \S\ref{sec:data-selection} and \S\ref{sec:training-setup}.

\item {\bf Open access to data and code}
    \item[] Question: Does the paper provide open access to the data and code, with sufficient instructions to faithfully reproduce the main experimental results, as described in supplemental material?
    \item[] Answer: \answerYes{} 
    \item[] Justification: We submit our data and code to facilitate reproduction of our results.

\item {\bf Experimental setting/details}
    \item[] Question: Does the paper specify all the training and test details (e.g., data splits, hyperparameters, how they were chosen, type of optimizer, etc.) necessary to understand the results?
    \item[] Answer: \answerYes{} 
    \item[] Justification: See \S\ref{sec:training}.

\item {\bf Experiment statistical significance}
    \item[] Question: Does the paper report error bars suitably and correctly defined or other appropriate information about the statistical significance of the experiments?
    \item[] Answer: \answerNo{} 
    \item[] Justification: We do not run our evaluations multiple times due to resource constraints, as these are LLM-based evaluations.

\item {\bf Experiments compute resources}
    \item[] Question: For each experiment, does the paper provide sufficient information on the computer resources (type of compute workers, memory, time of execution) needed to reproduce the experiments?
    \item[] Answer: \answerYes{} 
    \item[] Justification: See \ref{sec:training-setup}.
    
\item {\bf Code of ethics}
    \item[] Question: Does the research conducted in the paper conform, in every respect, with the NeurIPS Code of Ethics \url{https://neurips.cc/public/EthicsGuidelines}?
    \item[] Answer: \answerYes{} 
    \item[] Justification: Yes, we comply with these requirements.

\item {\bf Broader impacts}
    \item[] Question: Does the paper discuss both potential positive societal impacts and negative societal impacts of the work performed?
    \item[] Answer: \answerNA{} 
    \item[] Justification: This work does not have such societal impact.
    
\item {\bf Safeguards}
    \item[] Question: Does the paper describe safeguards that have been put in place for responsible release of data or models that have a high risk for misuse (e.g., pretrained language models, image generators, or scraped datasets)?
    \item[] Answer: \answerNo{} 
    \item[] Justification: This paper has no such risks.

\item {\bf Licenses for existing assets}
    \item[] Question: Are the creators or original owners of assets (e.g., code, data, models), used in the paper, properly credited and are the license and terms of use explicitly mentioned and properly respected?
    \item[] Answer: \answerYes{} 
    \item[] Justification: Yes, we provide proper citations.

\item {\bf New assets}
    \item[] Question: Are new assets introduced in the paper well documented and is the documentation provided alongside the assets?
    \item[] Answer: \answerYes{} 
    \item[] Justification: We provide detailed documentation on these assets.

\item {\bf Crowdsourcing and research with human subjects}
    \item[] Question: For crowdsourcing experiments and research with human subjects, does the paper include the full text of instructions given to participants and screenshots, if applicable, as well as details about compensation (if any)? 
    \item[] Answer: \answerNA{} 
    \item[] Justification: This work does not involve any hired human annotators. Annotations were done by co-authors.

\item {\bf Institutional review board (IRB) approvals or equivalent for research with human subjects}
    \item[] Question: Does the paper describe potential risks incurred by study participants, whether such risks were disclosed to the subjects, and whether Institutional Review Board (IRB) approvals (or an equivalent approval/review based on the requirements of your country or institution) were obtained?
    \item[] Answer: \answerYes{} 
    \item[] Justification: The project was reviewed by an IRB board and deemed exempt from federal regulations that govern human subject research.

\item {\bf Declaration of LLM usage}
    \item[] Question: Does the paper describe the usage of LLMs if it is an important, original, or non-standard component of the core methods in this research? Note that if the LLM is used only for writing, editing, or formatting purposes and does not impact the core methodology, scientific rigorousness, or originality of the research, declaration is not required.
    \item[] Answer: \answerNo{} 
    \item[] Justification: This work does not use LLMs in the described ways.

\end{enumerate}

\end{document}